\pdfoutput=1

\documentclass[11pt]{article}

\usepackage{EACL2023}

\usepackage{times}
\usepackage{latexsym}

\usepackage[T1]{fontenc}

\usepackage[utf8]{inputenc}

\usepackage{microtype}

\usepackage{inconsolata}

\usepackage{graphicx}
\usepackage{multirow}
\usepackage{tablefootnote}
\usepackage{footnote}
\usepackage{footmisc}
\usepackage{booktabs} 
\usepackage{comment} 
\usepackage{tikz}
\def\checkmark{\tikz\fill[scale=0.4](0,.35) -- (.25,0) -- (1,.7) -- (.25,.15) -- cycle;}
\usepackage{subfig}
\usepackage{linguex}
\newcommand{\minus}{\scalebox{0.75}[1.0]{$-$}}

\def\SPSB#1#2{\rlap{\textsuperscript{\textcolor{black}{#1}}}\SB{#2}}
\def\SB#1{\textsubscript{\textcolor{black}{#1}}}

\title{Why Can’t Discourse Parsing Generalize? \\ A Thorough Investigation of the Impact of Data Diversity} 

\author{Yang Janet Liu \and Amir Zeldes \\
        Department of Linguistics \\ Georgetown University \\ 
        {\tt \{yl879, amir.zeldes\}@georgetown.edu}}

\begin{document}
\maketitle

\begin{abstract}
  Recent advances in discourse parsing performance create the impression that, as in other NLP tasks, performance for high-resource languages such as English is finally becoming reliable. In this paper we demonstrate that this is not the case, and thoroughly investigate the impact of data diversity on RST parsing stability. We show that state-of-the-art architectures trained on the standard English newswire benchmark do not generalize well, even within the news domain. Using the two largest RST corpora of English with text from multiple genres, we quantify the impact of genre diversity in training data for achieving generalization to text types unseen during training. Our results show that a heterogeneous training regime is critical for stable and generalizable models, across parser architectures. We also provide error analyses of model outputs and out-of-domain performance. To our knowledge, this study is the first to fully evaluate cross-corpus RST parsing generalizability on complete trees, examine between-genre degradation within an RST corpus, and investigate the impact of genre diversity in training data composition.  
\end{abstract}

\section{Introduction}
\label{sec:introduction}

\begin{figure}
    \centering
    \includegraphics[width=0.9\columnwidth]{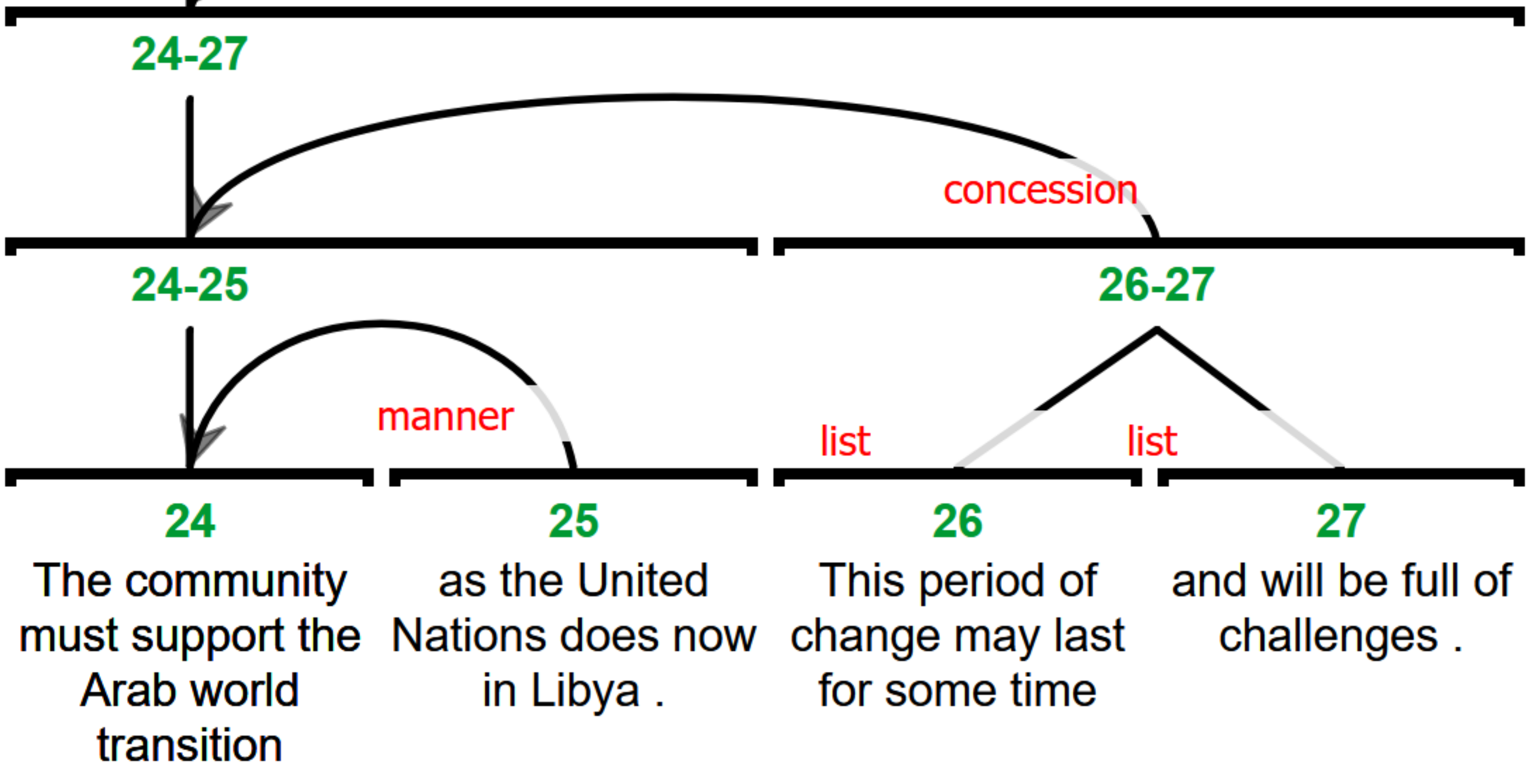}
    \caption{RST Fragment from GUM \cite{Zeldes2017}.}
    \vspace{-12pt}
    \label{fig:gum-rst-example}
\end{figure}

Discourse parsing is the task of identifying and classifying the coherence relations that hold between different parts of a text, such as recognizing that one sentence specifies the cause of events in another, or that a subordinate clause indicates the purpose of a main clause. In hierarchical frameworks, such as the Rhetorical Structure Theory (RST, \citealt{mann1988rhetorical}), parsers construct a labeled constituent tree of discourse units as shown in Figure \ref{fig:gum-rst-example}. Such trees have numerous applications: retrieving specific relations (e.g.~all \textsc{concessions} made in any speech by some politician, e.g.~unit 26-27 in Figure \ref{fig:gum-rst-example}), or hierarchically finding the most central unit in an arbitrary span of text (unit 24 in Figure \ref{fig:gum-rst-example}). They can also contribute to downstream tasks such as text generation \cite{maskharashvili-etal-2021-neural} and summarization (e.g.~\citealt{louis-etal-2010-discourse,li-etal-2016-role,xu-etal-2020-discourse,hewett-stede-2022-extractive}) and to qualitative analysis and comparison of texts \cite{wan-etal-2019-rst}.\footnote{This applies to other discourse frameworks too, such as PDTB \cite{PrasadWebberLeeEtAl2019} or SDRT \cite{asher2003logics}, but we limit the scope of this paper to RST parsing.}

Recent advances in NLP have resulted in increasingly accurate systems for tasks such as part-of-speech tagging \cite{heinzerling-strube-2019-sequence} and dependency parsing \cite{mrini-etal-2020-rethinking}, which are now commonly applied with confidence to novel data, at least for high-resource languages such as English. At the same time, rising scores on the main standard English RST benchmark, the RST Discourse Treebank (RST-DT, \citealt{CarlsonEtAl2003}), with data from the 1989 Wall Street Journal (WSJ), create the impression that discourse parsing too is becoming reliable, and by proxy, applicable to arbitrary text. This has resulted in use of automatic RST parsing to automatically generate large-scale datasets, such as MEGA-DT \cite{huber-carenini-2020-mega} or AMALGUM \cite{gessler-etal-2020-amalgum}. However, there are some indications that this picture is too optimistic, with some studies revealing problems in RST-DT-based parsing once the target domain changes (\citealt{ferracane-etal-2019-news,wang-etal-2019-using,Linscheid-et-al-rst,atwell-etal-2021-discourse}, \citeyear{atwell-etal-2022-change}; \citealt{nishida-matsumoto-2022-domain}; \citealt{yu-etal-2022-rst}), and qualitative inspection of parser outputs casting doubt on their reliability.


A problem in evaluating the stability of RST parsing for English is the scarcity of out-of-domain (OOD) data, as well as potential differences between the annotation schemes of different corpora.
Recently, the availability of increasing amounts of non-newswire data, as well as non-WSJ news data, in the English GUM corpus \cite{Zeldes2017} has made some detailed comparisons possible, which we explore below. Beyond evaluating OOD degradation, the existence of these large datasets for English RST allows us to test different joint training strategies to increase both the accuracy and generalizability of RST parsing for English.\footnote{Though we focus on English in this paper, understanding the size and nature of data needed in English will hopefully shed light on what other languages may require.} 

Our main goals here are: 1) to demonstrate the generalizability limitations of English RST parsing based on RST-DT and quantify them for users; 2) to explore reasons for generalizability issues, with a focus on the genre composition of training sets, pointing the way to the kind of data that robust discourse parsing requires; and 3) to promote multi-genre benchmarks for RST parsing based on our experimental results. 
Overall we find that
diverse training data leads to better generalization on unseen genres regardless of model architecture. 


\section{Related Work}
\label{sec:background}

\subsection{English RST Corpora}
\label{subsec:related-work-corpora}


The RST Discourse Treebank (RST-DT, \citealt{CarlsonEtAl2003}) is the standard English RST benchmark, with data from the WSJ section of the Penn Treebank (PTB, \citealt{MarcusSantoriniMarcinkiewicz1993}). Another human-annotated English RST corpus is the \textbf{G}eorgetown \textbf{U}niversity \textbf{M}ultilayer (GUM) corpus, which is freely available online and covers 12 written and spoken genres \cite{Zeldes2017}. GUM 
is continuously growing, with new data added in each version. For this paper, we used Version 8 of the corpus, which is described in Table \ref{tab:rst-eng-corpora} next to information about RST-DT (the current version of GUM, V9, was not yet available at submission time, and now contains 213 documents, 203K tokens and 26K EDUs, and the next version 10 is set to add 4 additional new genres).



\begin{table}[ht]
\centering
\resizebox{7cm}{!}{%
\begin{tabular}{l|cc}
\toprule
& \textbf{RST-DT} & \textbf{GUM V8} \\ \midrule
\textbf{documents} & 385 & 193 \\
\quad \textit{train/dev/test} & 347/$-$\tablefootnote{There is no established \texttt{dev} partition for RST-DT.}/38 & 145/24/24 \\
\textbf{tokens} & 203,352 & 180,851 \\
\textbf{EDUs} & 21,789 & 23,107 \\
\textbf{relation instances} & 20,163 & 21,903 \\
\textbf{relation labels} & 78 / 17 classes
& 32 / 15 classes \\
\textbf{genres} & 1\tablefootnote{Previous work has identified 4 sub-genres in the WSJ data \cite{webber-2009-genre}, but these are very unevenly distributed, and 29 RST-DT documents were not included in that analysis. } & 12 \\ \bottomrule
\end{tabular}
}
\caption{Overview of the English RST Corpora.}
\label{tab:rst-eng-corpora}
\vspace{-12pt}
\end{table}

As Table \ref{tab:rst-eng-corpora} shows, although GUM V8 is smaller than RST-DT in number of tokens and documents, it contains more instances of elementary discourse units (EDUs) and relations, since newswire units from RST-DT are longer on average. EDU segmentation guidelines are identical for the two corpora, and the label set is very similar, 
with both corpora using the \textsc{same-unit} pseudo-relation for discontinuous EDUs. 
GUM's 12 genres make it possible to conduct experiments to investigate parsing generalizability (see Table \ref{tab:gum8-data-overview} in Appendix \ref{sec:appendix:gum-content-breakdown} for their exact size breakdowns). Virtually all work on RST constituent parsing uses only the collapsed coarse relation classes (i.e.~17 labels for RST-DT); we will follow this practice for the evaluation below, but will refer to some issues relating to fine-grained labels in our analysis as well. In addition, we conducted an analysis of \textbf{s}atellite-\textbf{n}uclearity patterns in the two corpora, confirming that they are quite similar: NS dominates with 77.7\% in RST-DT, similar to 74.8\% in GUM news; interestingly, proportions in GUM overall are somewhat lower (70.1\%).

There are also other corpora annotated in RST dependencies, such as SciDTB \cite{yang-li-2018-scidtb} and full RST constituent treebanks in other languages (see \citealt{zeldes-etal-2021-disrpt} for an overview); since we focus on hierarchical English RST constituent parsing, we will use the corpora in Table \ref{tab:rst-eng-corpora} for the experiments below: they differ in content, vocabulary, domains, and more importantly, underlying communicative intents mirroring discourse relations, all of which are the backbone of our experiments to investigate English RST parsing generalizability as a function of training data.

\subsection{RST Discourse Parsers}
\label{subsec:parsers}

\begin{table*}[ht]%
\centering
\resizebox{\textwidth}{!}{%
\begin{tabular}{l|lllcllc}
\toprule
\multicolumn{1}{l|}{} &
  \multicolumn{1}{c}{\textbf{S}} &
  \multicolumn{1}{c}{\textbf{N}} &
  \multicolumn{1}{c|}{\textbf{R}} &
  \textbf{\begin{tabular}[c]{@{}c@{}}Neural\\ (Y/N)\end{tabular}} &
  \textbf{\begin{tabular}[c]{@{}l@{}}Model \\ Architectures\end{tabular}} & 
  \textbf{\begin{tabular}[c]{@{}l@{}}Additional \\ Features / Resources\end{tabular}} &
  \textbf{\begin{tabular}[c]{@{}c@{}}Pre-trained\\ LM\end{tabular}} \\ \midrule
\multicolumn{1}{c|}{} &
  \multicolumn{7}{c}{\textsc{bottom-up}} \\ \midrule
\multicolumn{1}{l|}{\citet{ji-eisenstein-2014-representation}$^{*}$} &
  64.1 &
  54.2 &
  \multicolumn{1}{l|}{46.8} &
  N &
  transition-based + SVMs &
  \begin{tabular}[c]{@{}l@{}}lexical, dependency, graphical features\end{tabular} &
  $-$ \\
\multicolumn{1}{l|}{\citet{yu-etal-2018-transition}$^\diamondsuit$} &
  71.4 &
  60.3 &
  \multicolumn{1}{l|}{49.2} &
  Y &
  \begin{tabular}[c]{@{}l@{}}transition-based + Bi-LSTM; \\ encoder-decoder + dynamic oracle\end{tabular} &
  implicit syntactic features &
  $-$ \\
\multicolumn{1}{l|}{\citet{guz-carenini-2020-coreference}$^\spadesuit$} &
  \textbf{76.5} &
  65.9 &
  \multicolumn{1}{l|}{\textbf{54.8}} &
  Y &
  \begin{tabular}[c]{@{}l@{}}transition-based + MLPs \end{tabular} &
  \begin{tabular}[c]{@{}l@{}}organizational features \cite{wang-etal-2017-two} \end{tabular} &
  SpanBERT-base \\ 
   \multicolumn{1}{l|}{\citet{yu-etal-2022-rst}$^\clubsuit$} &
  76.4 &
  \textbf{66.1} &
  \multicolumn{1}{l|}{54.5} &
  Y &
  \begin{tabular}[c]{@{}l@{}}\citet{yu-etal-2018-transition} + 2 pre-training tasks: Next EDU \\ Prediction; Discourse Marker Prediction \end{tabular} &
  $-$ &
  XLNet \\  \midrule
 &
 \multicolumn{7}{c}{\textsc{top-down}} \\ \midrule
 \multicolumn{1}{l|}{\citet{koto-etal-2021-top}$^\clubsuit$} &
  73.1 &
  62.3 &
  \multicolumn{1}{l|}{51.5} &
  Y &
  LSTM + dynamic oracle &
  \begin{tabular}[c]{@{}l@{}}sentence and paragraph boundaries \end{tabular} &
  $-$ \\
\multicolumn{1}{l|}{\citet{nguyen-etal-2021-rst}$^\clubsuit$} &
  74.3 &
  64.3 &
  \multicolumn{1}{l|}{51.6} &
  Y &
  \begin{tabular}[c]{@{}l@{}}Bi-LSTM encoder + \\ unidirectional LSTM decoder\end{tabular} &
  $-$ &
  XLNet-base \\
\multicolumn{1}{l|}{\citet{zhang-etal-2021-adversarial}$^\clubsuit$} &
  76.3 &
  \textbf{65.5} &
  \multicolumn{1}{l|}{\textbf{55.6}} &
  Y &
  \begin{tabular}[c]{@{}l@{}}split-point encoding and ranking + \\ adversarial learning\end{tabular} &
  $-$ &
  XLNet-base \\ 
\multicolumn{1}{l|}{\citet{liu-etal-2021-dmrst}$^\spadesuit$} &
  \textbf{76.5} &
  65.2 &
  \multicolumn{1}{l|}{54.2} &
  Y &
  pointer-network decoder + depth-first span splitting + biaffine classifiers &
  $-$ & XLM-RoBERTa-base \\
  \bottomrule
\end{tabular}%
}
\caption{Micro-averaged Parser Scores on RST-DT with Gold Segmentation. $^{*}$ = scores from \citet{morey-etal-2017-much}. $^\diamondsuit$ = converted Parseval scores for \citet{yu-etal-2018-transition}'s parser reported by \citet{koto-etal-2021-top}. $^\spadesuit$ = 5 run average Original Parseval replication of respective work. $^\clubsuit$ = scores from the original paper (not necessarily averaged scores). }%
\vspace{-8pt}
\label{tab:overview-rst-parsers}%
\end{table*}

Several approaches have been proposed for RST parsing, primarily distinguished by a \textsc{bottom-up} vs.~\textsc{top-down} approach, and the algorithm used (often a neural shift-reduce architecture). Table \ref{tab:overview-rst-parsers} compares recent high scoring parsers on three metrics: \textbf{S}pan (whether subtrees span the right EDUs), \textbf{N}uclearity (whether edges point the right way), and \textbf{R}elation (whether labels are correct). Following \citet{morey-etal-2017-much}, we use the more stringent original Parseval procedure on binary trees. 

As Table \ref{tab:overview-rst-parsers} shows, \textsc{bottom-up} and \textsc{top-down} approaches are currently very close, and all state-of-the-art (SOTA) systems rely on transformer word embeddings as the primary input. 
To rule out an impact of parser architecture, we test both the \textsc{bottom-up} from \citet{guz-carenini-2020-coreference}, using their best \texttt{SpanBERT-NoCoref} setting, and the \textsc{top-down} from \citet{liu-etal-2021-dmrst} as their code is public and both have near-SOTA S and N metrics, which are more robust to differences between label inventories across corpora (relevant hyperparameters and \texttt{dev} performance are provided in Appendix \ref{sec:appendix-model-details}).
\subsection{Cross-Genre Variation in RST}
\label{subsec:cross-genre-variation-in-rst}

Previous work suggests RST trees vary widely across text types,\footnote{We use ``text type'', ``genre'', and ``domain'' interchangeably here to mean different kinds of data, though they are not exactly the same (e.g.~see \citealt{biber1988} and \citealt{Lee2001GenresRT}).} in relation distributions \cite{taboada-and-lavid-2003,Zeldes2018book}, ways they are signaled (e.g.~\citealt{Eggins1997GenresAR,liu-2019-beyond,Demberg2019HowCA}), and variation across languages (\citealt{cao-etal-2018-rst}; \citealt{stede-neumann-2014-potsdam}; \citealt{Iruskieta2013TheRB}; \citealt{redeker-etal-2012-multi}; \citealt{da-cunha-etal-2011-development}; \citealt{russian-rst-corpus}). 

These differences suggest OOD parsing will be challenging: For example,
\citet{taboada-and-lavid-2003} showed that in appointment-scheduling conversations, \textsc{Concession}, \textsc{Condition}, \textsc{Cause}, and \textsc{Result} were disproportionately frequent due to the nature of the data, correlating with stages of a conversation (i.e.~Opening, Task Performance, Closing): e.g.~more \textsc{Restatement}, \textsc{Evaluation}, and \textsc{Summary} in the Closing stage. Similarly, \citet{Zeldes2018book} showed  \textsc{Motivation} often encodes how-to guides influencing one's willingness to act, which almost never occurs in the objective style of news. We therefore expect parsers trained on a single domain to generalize poorly to novel ones.



\subsection{Generalizability in RST Parsing}
\label{subsec:ood-rst}

Although some first studies are beginning to appear, generalizability in full RST parsing remains understudied. 
Three recent papers include some cross-dataset numbers, but not a systematic analysis of full RST constituent parser generalizability. \citet{atwell-etal-2021-discourse} examined the subtask of relation classification between PDTB \cite{PrasadWebberLeeEtAl2019} and RST annotated data, and included F1 scores for identifying relations in GUM V5 using the RST-DT training data for relations that appear in both datasets (however, GUM V5's EDU segmentation differed strongly from RST-DT and the parser in the paper was an older non-neural one). 

\citet{yu-etal-2022-rst} evaluate a neural parser on RST-DT and report scores on 11/12 genres in GUM V7 (same segmentation as RST-DT, but only 25 relations and less data than V8); however because their focus is not on cross-genre generalization, they do not report micro-averaged scores on the whole GUM corpus, do not evaluate training on GUM or joint training, and do not test held-out genres within GUM.
Finally, \citet{atwell-etal-2022-change} used GUM V6 (with RST-DT's segmentation and 8 genres) to identify relation types between pairs of provided argument spans, similar to PDTB-style shallow discourse parsing in the CoNLL'16 shared task \cite{xue-etal-2016-conll} and recent DISRPT shared task on relation classification \cite{zeldes-etal-2021-disrpt}.

This study is therefore the first to fully evaluate cross-genre RST parsing generalizability on complete trees in datasets with the same EDU segmentation. To our knowledge, no previous work has tested an OOD-trained parser on RST-DT, attempted joint training on both corpora, evaluated between-genre degradation within an RST corpus, or the impact of training data genre composition.


\section{Experiments}
\label{sec:experiments}

In this section we conduct four experiments. First, we want to find out how compatible the two English RST corpora are, and how well they generalize to each other's domains (\S\ref{subsec:cross-corpus}). Second, we want to test whether we can incorporate training data from GUM in order to exceed SOTA performance on RST-DT, with one of three approaches: joint training, model stacking, and pretraining (\S\ref{subsec:joint}). Third, we would like to know whether generalizability issues are mainly due to differences between corpora (details of annotation guidelines or common practices) or the nature of the texts' genres and domains themselves (\S\ref{subsec:GUM8-cross-genre-ova}). We test this by leaving out different subsets of GUM genres from training, and quantify OOD degradation for each genre, which also informs users of SOTA parsers of what degradation they can expect on genres `in the wild'. The final experiment (\S\ref{subsec:genre-variety-fixed-sized}) tests whether, given a fixed training set size, having relatively many `small' genres leads to better OOD stability than having few genres with more material each, when the target is a totally disjoint set of genres. The code, trained and finetuned models, and predicted GUM parses are available at \url{https://github.com/janetlauyeung/crossGENRE4RST}. 

\subsection{Cross-Corpus Generalization} 
\label{subsec:cross-corpus}


In this section we hypothesize that, since GUM contains many genres, models trained on it will degrade less when testing on RST-DT than in the opposite scenario. This is non-trivial, since the corpora are very close in total size, and it is possible that diverse training genres with little data each will prevent effective learning for any genre. We train the two parsers identified in \S\ref{subsec:parsers} from \citet{guz-carenini-2020-coreference} (\textsc{bottom-up}) and \citet{liu-etal-2021-dmrst} (\textsc{top-down}) on the \texttt{train} partition of each dataset, and report scores on the \texttt{test} set (since RST-DT has no \texttt{dev} partition, we follow previous work in using 10\% of training data stratified by the number of EDUs in each document as a \texttt{dev} set, which remains the same in the training). 



A possible limitation of our experiment is annotation differences between the corpora, especially for their relation inventories, which are similar, but not identical. We therefore omit R scores 
in Tables \ref{tab:cross-corpus-guz} and \ref{tab:cross-corpus-dmrst}, though we will return to relations in more detail below. For S and N, however, we expect differences to be small: both corpora use the same segmentation guidelines, and the same guideline of subordinating less prominent satellite units to more prominent nuclei using tests such as deletability and subjective prominence. Although EDU segmentation is not our focus, we can confirm the compatibility of segmentation guidelines by cross-testing the current SOTA EDU segmenter from \citet{gessler-etal-2021-discodisco}. The results show fairly modest degradation training on RST-DT and testing on GUM, from 94.9 to 89.9, while the opposite direction shows a 92.9 to 91 drop (see Table \ref{tab:cross-seg} in Appendix \ref{appendix:cross-corpus-seg-scores} for full numbers). We speculate that segmentation generalizes well due to the syntactic, and hence less sparse and domain-specific nature of the task, which resembles clause boundary detection. 



For cross-corpus S and N scores, both parsers show a very significant degradation when training on RST-DT to parse OOD data from GUM, as shown in Tables \ref{tab:cross-corpus-guz} and \ref{tab:cross-corpus-dmrst}. For instance, there is a degradation of ~$\sim$$\minus$11 points for S and $\sim$$\minus$16 for N using the the \textsc{bottom-up} architecture. We were also curious whether this applies to GUM's news genre too: indeed, testing only on GUM \textit{news} reduces degradation by about 50\% using the \textsc{bottom-up} parser, but still shows a substantial performance hit, even if the model is trained exclusively on news (albeit from much older, WSJ news data). By contrast, the GUM-trained model actually scores better on RST-DT than on GUM, with minor improvements of +2.8 on S and +0.4 on N using the \textsc{bottom-up} parser. The same degradation pattern is observed in the \textsc{top-down} parser performance: substantial degradation overall, and worse when training on news and testing other genres.\footnote{We further analyzed left vs.~right-branching trees during testing. In both corpora, NS outperformed SN: RST-DT in-domain scores are F1=82.6\% vs.~73.4\% for NS vs.~SN; GUM V8 has 81.6\% vs.~72.4\%, meaning SN transitions are ‘easier’.}


\begin{table}[h!tb]
\resizebox{\columnwidth}{!}{
\begin{tabular}{lllll}
\toprule
\textbf{train}  & \textbf{test}     & \textbf{S}                           & \textbf{N}                           & \textbf{R} \\
\midrule
RST-DT & RST-DT & 76.5  & 65.9 & 54.8       \\
\textit{}       & GUM      & 65.3 {\color[HTML]{FE0000} (-11.2)}  & 49.5 {\color[HTML]{FE0000} (-16.4)}  & --         \\
                & GUM \textit{news} & 71.0 {\color[HTML]{FE0000} (-5.5)} &  57.5 {\color[HTML]{FE0000}(-8.4)} & --         \\
                \midrule
GUM   & GUM     & 69.9  & 57.0 & 48.5      \\
\textit{}       & RST-DT  &  72.7 {\color[HTML]{3531FF} (+2.8)}   & 57.4 {\color[HTML]{3531FF} (+0.4)}   & -- \\
                & GUM \textit{news} & 71.6 &  58.5 & 49.5 \\ 
\bottomrule
\end{tabular}
}
\vspace{-5pt}
\caption{Cross-Corpus Results (5 run average) of the \textsc{bottom-up} Parser from \citet{guz-carenini-2020-coreference}.}
\label{tab:cross-corpus-guz}
\end{table}

\begin{table}[h!tb]
\resizebox{\columnwidth}{!}{
\begin{tabular}{lllll}
\toprule
\textbf{train}  & \textbf{test}  & \textbf{S}  & \textbf{N}  & \textbf{R} \\
\midrule
RST-DT & RST-DT & 76.5  & 65.2 & 54.2       \\
\textit{}       & GUM      & 66.2 {\color[HTML]{FE0000} (-10.3)}  & 50.8 {\color[HTML]{FE0000} (-14.4)}  & --         \\
& GUM \textit{news} & 67.9 {\color[HTML]{FE0000} (-8.6)} &  55.8 {\color[HTML]{FE0000}(-9.4)} & --         \\\midrule
GUM   & GUM     & 68.6  & 54.9 & 46.1      \\
\textit{}       & RST-DT  &  71.1 {\color[HTML]{3531FF} (+2.5)}   & 55.9 {\color[HTML]{3531FF} (+1.0)}   & -- \\
                & GUM \textit{news} & 73.4 &  63.3 & 57.2 \\ 
\bottomrule
\end{tabular}
}
\vspace{-5pt}
\caption{Cross-Corpus Results (5 run average) of the \textsc{top-down} Parser from \citet{liu-etal-2021-dmrst}.}
\vspace{-8pt}
\label{tab:cross-corpus-dmrst}
\end{table}

We interpret this result to mean that unlike in EDU segmentation, where results are close in both directions, genre composition of the \texttt{train} and \texttt{test} data plays a crucial role in the generalizability of RST constituent parsing, regardless of parser architecture. Thus, we opt for the \textsc{bottom-up} parser for the experiments presented in the rest of the paper as it outperformed the \textsc{top-down} parser on both corpora. 

To be clear, the results do not suggest that training on GUM is a way of achieving top performance on RST-DT: the GUM model's scores are still almost 4 and 8 points below the RST-DT trained model for S and N. However, it seems that RST-DT news data is less surprising for the GUM model which has already seen some news, and in sum, RST-DT data appears to be a comparatively `easy' target given the broad genre inventory that the GUM model is trained to tackle. 

These results also confirm that training on multiple genres, each with comparatively fewer documents, can lead to good performance with only minor degradation on the very narrow WSJ domain from RST-DT. While we think this result by itself is important and suggests that RST parsing work should devote more attention to multi-genre corpora as benchmarks, it leaves a question open: Can we combine data from both corpora to boost English RST parsing performance on RST-DT?

\subsection{Joint Training}\label{subsec:joint}

If RST parsing generalizes across domains, it may be possible to see gains through joint training. For this we compare three approaches: naive concatenation, model stacking, and pretraining, which we evaluate on the RST-DT benchmark. 



For concatenation (\textsc{concat}) we must map relations across corpora. As in previous work, we target coarse relation classes, most of which are identical in both corpora. Exceptions include GUM's \textsc{phatic} relation used for dysfluencies and back-channeling (due to conversational data), GUM's \textsc{preparation}, and the mapping between RST-DT's \textsc{Temporal} and \textsc{Comparison} classes, which map differently onto GUM's \textsc{circumstance} and \textsc{sequence}, or GUM's \textsc{antithesis}, depending on the precise senses. Although the number of relation instances affected by these mismatches is modest, it is not negligible (13.3\%), and we do not expect this approach to outperform in-domain training, and mainly report on it for completeness. Our mapping is given in Appendix \ref{sec:appendix-relation-mapping}. 

For model stacking, we test three variants: 1) \textsc{flair-label}: train an LSTM using FLAIR \cite{akbik-etal-2019-flair} to predict EDU dependency labels: the LSTM receives text for three-EDU chunks, set apart by separators, and predicts the middle EDU's label in the GUM corpus scheme (using the RST dependency conversion from \citealt{li-etal-2014-text}). We then train the parser on RST-DT with predicted GUM labels for each EDU as an additional feature encoded as a dense embedding, requiring no relation mapping. 2) \textsc{sr-label}: train a full shift-reduce parser on GUM, generate predictions for RST-DT in the GUM scheme, and collapse these into dependencies to create the same kind of features; this approach gives the GUM classifier more access to global context than the previous LSTM's EDU triples, but may be more vulnerable to sparseness. 3)  \textsc{sr-graph}: because it is possible that incompatibility of RST-DT and GUM labels may cause confusion, we also attempted using the same parser as in the previous approach, but featurizing each EDU's predicted dependency attachment direction, and EDU distance to the parent EDU, instead of the label itself. Finally, we test a simple pretraining approach, in which we finetune the underlying SpanBERT model \cite{joshi-etal-2020-spanbert} on full parsing of the GUM corpus, then load the fine-tuned SpanBERT, and train again on RST-DT (\textsc{sr-ft}). 


Table \ref{tab:rstdt-categorical} shows that \textsc{sr-ft} achieved the best performance compared to the other approaches and is on par with the pure in-domain training on RST-DT for S and R compared to previous work. There is a minor but stable gain on average (65.9\textrightarrow66.2) for N in \textsc{sr-ft}, which was verified by rerunning the experiment, as well as the selected SOTA system from \citet{guz-carenini-2020-coreference}, 5 times. The remaining scenarios are virtually equivalent to training on RST-DT alone, suggesting that added features are more distracting than helpful. 

\begin{table}[htb]
\centering
\resizebox{\columnwidth}{!}{%
\begin{tabular}{l|ccc|cc}
\toprule
 & \textbf{S} & \textbf{N} & \textbf{R} & \textit{architecture} \\ \midrule
\citet{zhang-etal-2021-adversarial}$^{*}$ & 76.3 & 65.5 & \textbf{55.6} & \textsc{top-down} \\
\citet{liu-etal-2021-dmrst}$^\diamondsuit$ & \textbf{76.5} & 65.2 & 54.2 & \textsc{top-down} \\
\citet{guz-carenini-2020-coreference}$^\diamondsuit$ & \textbf{76.5} & \textbf{65.9} & 54.8 & \textsc{bottom-up} \\
\midrule
\textit{this paper (\textsc{concat})}$^\spadesuit$ & 75.9 & 64.8 & 54.1 & \multirow{5}{*}{\textsc{bottom-up}} \\ 
\textit{this paper (\textsc{flair-label})}$^\spadesuit$ & 75.8 & 65.6 & 55.3 &  \\ 
\textit{this paper (\textsc{sr-label})}$^\spadesuit$ & 76.2 & 66.0 & 55.3 &  \\ 
\textit{this paper (\textsc{sr-graph})}$^\spadesuit$ & 75.8 & 65.5 & 54.7 \\ 
\textit{this paper (\textsc{sr-ft})}$^\diamondsuit$ & \textbf{76.3} & \textbf{66.2} & \textbf{55.5} \\ \midrule
Human \cite{morey-etal-2017-much} & 78.7 & 66.8 & 57.1 & $-$ \\
\bottomrule
\end{tabular}%
}
\vspace{-3pt}
\caption{Joint Training Performance on RST-DT. $^{*}$ = original paper score. $^\diamondsuit$ = 5 run avg.; $^\spadesuit$ = 3 run avg. }
\vspace{-10pt}
\label{tab:rstdt-categorical} 
\end{table}

This result is somewhat surprising given that scores are not very high, and there should still be headroom for improvement; however, we suspect some of the missing information responsible for errors may relate to global structure and pragmatic understanding which cannot easily be compensated for by adding more genres with potentially disjoint vocabulary. 
A further surprising result is that \textsc{concat} is not much worse than the base system, suggesting that most of the score comes from obvious cases (e.g.~relative and infinitival clauses) which do not differ substantially across corpora or genres, and were already learned without the added data. Similar results have been found for other NLP tasks where adding a second dataset for joint training creates a `break-even' effect: the benefit of more data helps about as much as the disparate domains harm within-corpus performance, e.g.~for dependency parsing in Hebrew \cite{zeldes-etal-2022-second} and English \cite{ZeldesSchneider2023UDreport}, English coreference resolution \cite{zhu-etal-2021-ontogum}, and very recent similar results for RST parsing for Chinese \cite{peng-etal-2022-gcdt}.

\subsection{OOD Multi-Genre Degradation} 
\label{subsec:GUM8-cross-genre-ova}

In this section we explore our next question: How badly will a multi-genre trained model degrade on unseen genres, when the annotation scheme remains identical? This question is important for applied, methodological, and ethical reasons. From the practical perspective, while we already know that training only on news leads to severe degradation, we want to inform users of discourse parsing about expected performance on unseen domains if training data is already domain-rich. Methodologically, we want to test the benefit of adding multiple genres and weigh the differences between a few-genre corpus design and a many-genres design, given a fixed total data size capacity for dataset creation. Finally from an ethical perspective, OOD degradation has real life implications for less common types of data, whether they come from underrepresented genres, communicative situations, or speaker demographics \cite{MengeshaEtAl2021}. While we cannot fix these problems without more data, we can point them out and increase awareness of skewed data biases at the discourse level.

To explore OOD degradation, we conducted 10 experiments, comparing the normal genre-balanced scenario (GUM \texttt{test}) with testing on each genre when it is not in `train' (\textsc{One-vs-All} or \textsc{ova}). For consistency, we test \textsc{ova} for the 8 roughly equal sized non-growing genres in GUM. Since data for the smaller 4 growing genres may be less reliable and non-comparable, we separately report scores for training on all 8 large genres (\textsc{all-large}), tested on each of the four growing genres. 

In the last scenario in particular, we would like to see whether small test genres, for which we cannot obtain enough training data, perform better if a training genre exists which may offer a near substitute. For instance, the small \textit{textbook} and \textit{speech} genres have structural organization and formal language similar to \textit{academic} and \textit{news} respectively, and there is some overlap between \textit{speech} vocabulary and \textit{interviews} with politicians. By contrast,  \textit{vlog} and \textit{conversation} are highly informal and colloquial, perhaps closest to the \textit{reddit} or \textit{interview} genres, but still likely more challenging. 
For reproducibility, exact training data compositions are given in Table \ref{tab:ova-data-description} in Appendix \ref{sec:apendix-data-overview-ova}. 

Table \ref{tab:gum-ova-test-degradation} shows the results, with the same 4 documents from each held-out genre used to evaluate each \textsc{ova} model (these are not included in any training set for consistency). The \textit{degradation} column shows that the parser suffers when a genre is removed from training across the board, except for \textit{news} and the Span level of \textit{reddit}; \textit{bio} and \textit{how-to} genres suffer the most: qualitative inspection reveals that this is due to frequent errors on \textsc{Organization} and \textsc{Explanation}, which tend to lack explicit cues and require understanding of macro-structure and global topic or pragmatic reasoning to interpret. It is likely that lack of a similar substitute to these genres contributes to the degradation. 


By contrast, degradation on \textit{interview} is low (perhaps covered by \textit{conversation} and \textit{news} in training), and the inverse result for \textit{news} suggests that collecting more news data may not be a priority. The minimal degradation for \textit{reddit} is surprising, and is likely due to more instances of explicitly signaled and `easy' relations such as \textsc{Purpose} (e.g.~\textit{in order to}) and \textsc{Contingency} (e.g.~\textit{if}), but we note that ordinary scores on \textit{reddit} are low to begin with (second worst after \textit{fiction}, which degrades more). 
The bottom of the table shows smaller genres suffering, except for a minor gain in \textit{speech} at the Span level. It is unsurprising that \textit{conversation} degrades the most, especially at the R level, since it contains challenging phenomena absent in other genres, such as back-channeling, dysfluencies, and abrupt topic changes. This is mirrored in bad performance on the \textsc{Organization} and \textsc{Topic} relation classes (see more detailed analysis in \S\ref{sec:error-analysis}). 

\begin{table}[htb]
\centering
\resizebox{7cm}{!}{%
\begin{tabular}{l|rrr|rrr|rrr}
\toprule
& \multicolumn{3}{c|}{\textbf{GUM \texttt{test}}} & \multicolumn{3}{c|}{\textbf{ova}} & \multicolumn{3}{c}{\textit{degradation}} \\ \midrule
\textbf{non-growing} &
\multicolumn{1}{c}{\textbf{S}} &
\multicolumn{1}{c}{\textbf{N}} &
\multicolumn{1}{c|}{\textbf{R}} &
\multicolumn{1}{c}{\textbf{S}} &
\multicolumn{1}{c}{\textbf{N}} &
\multicolumn{1}{c|}{\textbf{R}} &
\multicolumn{1}{c}{\textbf{S}} &
\multicolumn{1}{c}{\textbf{N}} &
\multicolumn{1}{c}{\textbf{R}} \\ \midrule
\textit{academic}  & 77.0 & 68.5 & 59.8 & 75.2 & 66.2 & 55.7 & 1.7  & 2.3  & 4.1  \\
\textit{bio}       & 70.4 & 58.2 & 51.2 & 68.8 & 53.9 & 43.2 & 1.6  & 4.3  & \textbf{8.0}  \\
\textit{fiction}   & 66.3 & 53.1 & 43.7 & 64.5 & 50.1 & 42.1 & 1.8  & 3.0  & 1.7  \\
\textit{interview} & 73.3 & 59.0 & 50.9 & 73.0 & 56.7 & 49.7 & 0.3  & 2.2  & 1.2  \\
\textit{news}      & 71.7 & 58.4 & 49.1 & 72.2 & 59.2 & 51.3 & -0.5 & -0.8 & -2.2 \\
\textit{reddit}    & 66.0 & 52.3 & 44.2 & 66.6 & 51.9 & 43.3 & 0.6 & 0.4  & 0.8  \\
\textit{voyage}    & 78.3 & 62.1 & 51.8 & 77.4 & 59.7 & 49.3 & 0.9  & 2.4  & 2.4  \\
\textit{how-to}    & 76.5 & 63.6 & 54.6 & 67.1 & 54.3 & 44.8 & \textbf{9.3}  & \textbf{9.3}  & \textbf{9.9}  \\ \midrule
& \multicolumn{3}{c|}{\textbf{GUM \texttt{test}}} & \multicolumn{3}{c|}{\textbf{\textsc{all-large}}} & \multicolumn{3}{c}{\textit{degradation}} \\ \midrule
\textbf{growing} &
\multicolumn{1}{c}{\textbf{S}} &
\multicolumn{1}{c}{\textbf{N}} &
\multicolumn{1}{c|}{\textbf{R}} &
\multicolumn{1}{c}{\textbf{S}} &
\multicolumn{1}{c}{\textbf{N}} &
\multicolumn{1}{c|}{\textbf{R}} &
\multicolumn{1}{c}{\textbf{S}} &
\multicolumn{1}{c}{\textbf{N}} &
\multicolumn{1}{c}{\textbf{R}} \\ \midrule
\textit{conversation} & 45.4 & 34.5 & 26.7 & 42.7 & 31.4 & 21.8 & 2.7  & 3.1 & \textbf{4.9} \\
\textit{speech}       & 76.0 & 64.4 & 55.2 & 76.4 & 62.9 & 54.8 & -0.4 & 1.5 & 0.4 \\
\textit{textbook}     & 77.4 & 66.8 & 57.3 & 76.2 & 64.3 & 54.5 & 1.2  & 2.6 & 2.9 \\
\textit{vlog}         & 64.8 & 49.0 & 42.8 & 63.3 & 49.0 & 40.4 & 1.5  & 0.0 & 2.5 \\ \bottomrule
\end{tabular}%
}
\caption{Per Genre Scores for GUM \texttt{test} vs.~the \textsc{ova} or \textsc{all-large} Experiments (3 run average).}
\vspace{-8pt}
\label{tab:gum-ova-test-degradation}
\end{table}

\subsection{Genre Variety in a Fixed-Size Sample}
\label{subsec:genre-variety-fixed-sized}

While \S\ref{subsec:GUM8-cross-genre-ova} shows how genres differ in degradation, it falls short of proving that genre diversity promotes generalization when all other things are equal, since train sets for each genre are not identical in size. Our final experiment addresses this: ideally, we want to compare scores on a fixed OOD test set for equal-sized training corpora, divided into fewer or more genres. Although we might expect more genres to be helpful for generalization, this is not trivial or inevitable: If there are not enough recurring examples of infrequent phenomena, because data is so diverse, learning might fail due to sparseness; that is, more genres could be distracting rather than helpful in a meaningful way, which could hurt performance.

Because genres in GUM are small ($\sim$2K EDUs) and we want a critical mass of 5K+ EDUs for reasonable parser performance, our combinatory options are limited to 3+ training genres. The parser can only be trained on complete documents (we cannot select \textit{n} EDUs from each genre $-$ genres must comprise coherent texts), i.e.~we must find permutations of the data which total roughly the same number of training instances, but with different genres. Table \ref{tab:fixed-size} gives details on the 3 best training cohorts based on these criteria.

\begin{table}[ht]
\centering
\resizebox{\columnwidth}{!}{%
\begin{tabular}{llcl|llcl}
\toprule
\textbf{ID} & \textbf{genres}     & \textbf{docs} & \textbf{EDUs} & \textbf{ID} & \textbf{genres}     & \textbf{docs} & \textbf{EDUs} \\
\midrule
C1           & \textit{academic}  & 18            & 1,970         & C3           & \textit{academic}  & 9            & 1,004         \\
\textit{}   & \textit{bio}       & 19            & 1,981         & \textit{}   & \textit{bio}       & 9            & \phantom{$-$}930           \\
\textit{}   & \textit{news}      & 23            & 1,760         & \textit{}   & \textit{news}      & 10            & \phantom{$-$}635           \\
\cline{1-4}
            & \textbf{total}     &      60         & 5,711         &             &                    &               &               \\
\cline{1-4}
C2           & \textit{fiction}   & 15            & 1,941         &             & \textit{fiction}   & 8            & 1,027         \\
\textit{}   & \textit{interview} &   15            & 1,931         &             & \textit{interview} &    8           & 1,199         \\
            & \textit{how-to}    &  15             & 1,840         &             & \textit{how-to}    &       8        & \phantom{$-$}917           \\
            \midrule
\textit{}   & \textbf{total}     &       45        & 5,712         &             & \textbf{total}     &       52        & 5,712         \\
\bottomrule
\end{tabular}
}
\caption{Composition of 3 Fixed-Size Training Cohorts with Different Genre Contents.}
\vspace{-5pt}
\label{tab:fixed-size}
\end{table}

\begin{table*}[ht]
\centering
\resizebox{15cm}{!}{%
\begin{tabular}{l|rrr|l|rrr|l|rrr|l|rrr|l|rrr|l|rrr}
\toprule
  &
  \multicolumn{3}{c|}{\textbf{C1}} &
  \multirow{10}{*}{\textbf{}} &
  \multicolumn{3}{c|}{\textbf{C2}} &
  \multirow{10}{*}{\textbf{}} &
  \multicolumn{3}{c|}{\textbf{C3}} &
  \multirow{10}{*}{\textbf{}} &
  \multicolumn{3}{c|}{\textbf{C3$-$C1}} &
  \multirow{10}{*}{\textbf{}} &
  \multicolumn{3}{c|}{\textbf{C3$-$C2}} &
  \multirow{10}{*}{\textbf{}} &
  \multicolumn{3}{c}{\textbf{mean\_C3\_gain}} \\ \cline{1-4} \cline{6-8} \cline{10-12} \cline{14-16} \cline{18-20} \cline{22-24} 
 \textbf{test} & 
  \multicolumn{1}{c}{\textbf{S}} &
  \multicolumn{1}{c}{\textbf{N}} &
  \multicolumn{1}{c|}{\textbf{R}} &
   &
  \multicolumn{1}{c}{\textbf{S}} &
  \multicolumn{1}{c}{\textbf{N}} &
  \multicolumn{1}{c|}{\textbf{R}} &
   &
  \multicolumn{1}{c}{\textbf{S}} &
  \multicolumn{1}{c}{\textbf{N}} &
  \multicolumn{1}{c|}{\textbf{R}} &
   &
  \multicolumn{1}{c}{\textbf{S}} &
  \multicolumn{1}{c}{\textbf{N}} &
  \multicolumn{1}{c|}{\textbf{R}} &
   &
  \multicolumn{1}{c}{\textbf{S}} &
  \multicolumn{1}{c}{\textbf{N}} &
  \multicolumn{1}{c|}{\textbf{R}} &
   &
  \multicolumn{1}{c}{\textbf{S}} &
  \multicolumn{1}{c}{\textbf{N}} &
  \multicolumn{1}{c}{\textbf{R}} \\ \cline{1-4} \cline{6-8} \cline{10-12} \cline{14-16} \cline{18-20} \cline{22-24} 
\textit{conversation} &
  34.8 &
  23.4 &
  13.9 &
   &
  40.3 &
  27.9 &
  18.0 &
   &
  37.9 &
  26.4 &
  18.0 &
   &
  3.0 &
  3.0 &
  4.1 &
   &
  -2.5 &
  -1.5 &
  0.0 &
   &
  0.3 &
  0.7 &
  2.0 \\
\textit{reddit} &
  60.3 &
  45.3 &
  36.0 &
   &
  63.5 &
  46.9 &
  37.6 &
   &
  61.8 &
  47.6 &
  37.3 &
   &
  1.5 &
  2.3 &
  1.4 &
   &
  -1.7 &
  0.7 &
  -0.3 &
   &
  -0.1 &
  1.5 &
  0.6 \\
\textit{speech} &
  72.5 &
  58.2 &
  46.9 &
   &
  72.6 &
  59.3 &
  47.7 &
   &
  71.6 &
  57.1 &
  48.0 &
   &
  -0.9 &
  -1.1 &
  1.1 &
   &
  -1.0 &
  -2.1 &
  0.3 &
   &
  -0.9 &
  -1.6 &
  0.7 \\
\textit{textbook} &
  73.6 &
  59.0 &
  48.9 &
   &
  70.9 &
  55.0 &
  45.6 &
   &
  74.0 &
  60.5 &
  51.4 &
   &
  0.5 &
  1.5 &
  2.5 &
   &
  3.1 &
  5.5 &
  5.9 &
   &
  1.8 &
  3.5 &
  4.2 \\
\textit{vlog} &
  57.8 &
  41.3 &
  35.0 &
   &
  58.8 &
  44.5 &
  35.3 &
   &
  57.7 &
  43.4 &
  34.8 &
   &
  -0.1 &
  2.1 &
  -0.2 &
   &
  -1.1 &
  -1.1 &
  -0.5 &
   &
  -0.6 &
  0.5 &
  -0.3 \\
\textit{voyage} &
  76.6 &
  58.1 &
  47.5 &
   &
  76.5 &
  57.4 &
  46.4 &
   &
  78.0 &
  59.1 &
  50.2 &
   &
  1.5 &
  1.0 &
  2.7 &
   &
  1.6 &
  1.7 &
  3.8 &
   &
  1.5 &
  1.4 &
  3.3 \\ \cline{1-4} \cline{6-8} \cline{10-12} \cline{14-16} \cline{18-20} \cline{22-24} 
macro\_avg &
  62.6 &
  47.6 &
  38.0 &
   &
  \textbf{63.8} &
  48.5 &
  38.4 &
   &
  63.5 &
  \textbf{49.0} &
  \textbf{40.0} &
   &
  0.9 &
  1.5 &
  1.9 &
   &
  -0.3 &
  0.5 &
  1.5 &
   &
  0.3 &
  1.0 &
  1.7 \\
micro\_avg &
  58.7 &
  44.2 &
  34.8 &
   &
  \textbf{60.5} &
  45.7 &
  35.7 &
   &
  59.8 &
  \textbf{45.9} &
  \textbf{36.9} &
   &
  1.1 &
  1.7 &
  2.1 &
   &
  -0.6 &
  0.2 &
  1.2 &
   &
  0.2 &
  1.0 &
  1.6 \\ \bottomrule
\end{tabular}%
}
\vspace{-5pt}
\caption{Performance of 3 Fixed-Size Train Cohorts
with Different Genre Contents (5 run average). }
\vspace{-10pt}
\label{tab:57k-edu-results}
\end{table*}

As Table \ref{tab:fixed-size} shows, the greatest care was taken to ensure that the cohorts sum up to almost exactly as many training instances ($\sim$5,712 EDUs), at the price of somewhat diverse amounts of EDUs and documents per genre.  If we cannot control all factors, we prefer to constrain the amount of shift-reduce operations learned, in order to prevent any alternative explanation in which total size effects outweigh genre diversity. Additionally, we assume that EDU count per genre will vary in any multi-genre corpus due to complete document constraints; corpus developers are more likely to target a total fixed size budget. If having too many small genres is harmful, we expect cohort 3 (C3) to perform worst; by contrast, if diversity is helpful, C3 should perform best. In either case, we compare how the S/N/R metrics behave in each training regime.

Table \ref{tab:57k-edu-results} shows 5-run averages on each OOD genre, as well as total micro- and across-genres macro-averaged performance. The 3-genre cohorts, C1 and C2, perform similarly overall, though C2 outperforms C1 on test data (which is held constant for all cohorts). R scores are very close on macro-average, but C1 is particularly bad on \textit{conversation}, where C2 is tied with C3 on R (possibly because both have \textit{interview} data). Conversely, C1 is better on textbooks and travel guides (\textit{voyage}) than C2 (possibly because C1 has \textit{academic}, which predicts textbooks reasonably well, and bios -- a descriptive informational genre, something missing from C2). 

Overall best performance is obtained by C3, with 6 training genres, but the least data in each. Although C3 loses to C2 in the dialogue-oriented \textit{conversation} and \textit{reddit} genres, degradation is very modest and does not affect all metrics (e.g.~C3 has a better N score on \textit{reddit}).
C3 also outperforms C1 on \textit{academic} and \textit{voyage}, giving the best performance on both. The average gain for C3 across all metrics (3-metric average) is around +1\%---taking individual scores from each of five runs, and each genre test set as a data-point, this improvement is significant at $p<0.05$. For individual metrics, we see $\sim$+1.7\% on R, +1\% on N, and just under +0.3\% for S (significant at $p<0.05$ except for S).

Although all scores are rather low due to the small corpus sizes (about \textonequarter~of GUM), they suggest that more training genres with smaller portions each promotes OOD generalization, though not by a lot. It is an open question whether this gap would increase or decrease with corpus size: on the one hand, more data would allow for more lexical diversity even with few genres. On the other, it is likely that scores in small data are driven by easy to learn cases (e.g.~relative clauses as \textsc{Elaboration}, or \textsc{Purpose} infinitives), which stand to gain less from diversity. If more data means models will tackle more sparse phenomena, then genre diversity should matter \textit{more} for OOD material as the training set grows. To an extent, results in \S\ref{subsec:cross-corpus} showing worse generalization from the large but homogeneous RST-DT to GUM seem to support this hypothesis. We interpret the present results to mean that development of more diverse multi-genre data should take priority over building up material in existing genres to promote generalizable parsing.

\section{Error Analysis}
\label{sec:error-analysis}

Due to space, we limit our error analysis to examining dependency conversions of gold vs.~predicted trees, which allow us to break down OOD errors by coarse relation class in Table \ref{tab:errors} from all test sets in \S\ref{subsec:GUM8-cross-genre-ova}. We select models  with scores closest to average run scores. It is clear that even in the more modest degradation within GUM, over half of relation classes have $<50\%$ accuracy, with the document \textsc{root} being the hardest to identify---i.e.~the Central Discourse Unit (CDU)---followed by \textsc{Restatement} and \textsc{Evaluation}, which require reasoning over many EDUs and lack consistent overt signals, as opposed to relations such as \textsc{Attribution} (marked by speech verbs), \textsc{Purpose} (marked by purpose infinitives or \textit{in order to}), and \textsc{Contingency} (usually \textit{if}); interrupted clauses (\textsc{same-unit}) are easier as well. For CDU identification \cite{IruskietaEtAl2016}, which can benefit summarization or long-form QA systems, half of the genres (\textit{academic}, \textit{fiction}, \textit{interview}, \textit{voyage}, \textit{how-to}, \textit{vlog}) score 0\%; the highest accuracy is only 50\% (\textit{bio}, \textit{news}, \textit{reddit} and \textit{speech}).  
More alarmingly, in the cross-corpus setting (\S\ref{subsec:cross-corpus}), an RST-DT trained model captures only a single GUM CDU correctly (\textsc{acc}=0.042 vs.~0.375 for a GUM-trained model); scores on RST-DT are much higher: \textsc{acc}=0.842 for \textsc{sr-ft} trained on RST-DT vs.~0.553 for a GUM-trained model.

\begin{table}[h!tb]
\centering
\resizebox{7cm}{!}{%
\begin{tabular}{ll|ll}
\toprule
\textbf{\begin{tabular}[c]{@{}l@{}}gold coarse \\ relation class\end{tabular}} &
  \multicolumn{1}{l|}{\textbf{acc}} &
  \textbf{\begin{tabular}[c]{@{}l@{}}gold coarse \\ relation class\end{tabular}} &
  \multicolumn{1}{l}{\textbf{acc}} \\ \midrule
\textsc{Attribution} & 0.875 & \textsc{Context}      & 0.471        \\
\textsc{Purpose}     & 0.861 & \textsc{Adversative}  & 0.467        \\
\textsc{same-unit}   & 0.814 & \textsc{Organization} & 0.463        \\
\textsc{Contingency} & 0.794 & \textsc{Explanation}  & 0.431        \\
\textsc{Elaboration} & 0.666 & \textsc{Causal}       & 0.384        \\
\textsc{Joint}       & 0.654 & \textsc{Evaluation}   & 0.362        \\
\textsc{Topic}       & 0.574 & \textsc{Restatement}  & 0.308        \\
\textsc{Mode}        & 0.504 & \textsc{root}         & 0.208        \\ \bottomrule
\end{tabular}%
}
\caption{OOD Accuracy by Relation Class in GUM. }
\label{tab:errors}
\vspace{-10pt}
\end{table}

\begin{table}[h!tb]
\centering
\resizebox{6.5cm}{!}{%
\begin{tabular}{llr}
\toprule
\textbf{genre} & \textbf{gold coarse relation class} & \textbf{abs(resid)} \\
\midrule
\textit{textbook}       & \textsc{Context}             & 3.64          \\
\textit{speech}         & \textsc{Explanation}         & 3.14          \\
\textit{reddit}         & \textsc{Explanation}         & 3.02          \\
\textit{fiction}        & \textsc{Evaluation}          & 2.59          \\
\textit{bio}            & \textsc{Causal}              & 2.26          \\
\textit{vlog}           & \textsc{Causal}              & 2.23          \\
\textit{conversation}   & \textsc{Organization}        & 2.14         \\
\textit{voyage}         & \textsc{Context}             & 2.13          \\
\textit{academic}       & \textsc{Organization}        & 1.84          \\
\textit{how-to}         & \textsc{Organization}        & 1.62          \\
\textit{news}           & \textsc{Explanation}         & 1.38          \\
\textit{interview}      & \textsc{Evaluation}          & 0.89          \\ \bottomrule
\end{tabular}
}
\vspace{-3pt}
\caption{Max Absolute Error Residuals by Genre.}
\label{tab:err-residuals}
\vspace{-10pt}
\end{table}

  

	

Errors are also skewed by genre: Table \ref{tab:err-residuals} gives the gold label most surprisingly associated with errors per genre, given global error distributions, based on absolute $\chi^2$ residuals. While \textsc{Evaluation} is problematic in \textit{fiction} and \textit{interview}, \textsc{Explanation} and \textsc{Organization} are surprisingly hard to predict in 3 genres each: the former is used e.g.~in speeches and news for supporting evidence or justifications which are difficult to identify using lexical items, while the latter is used in conversations for phatic responses and back-channeling.  





\begin{figure}[htp]
    \centering
    \includegraphics[width=69mm,scale=1]{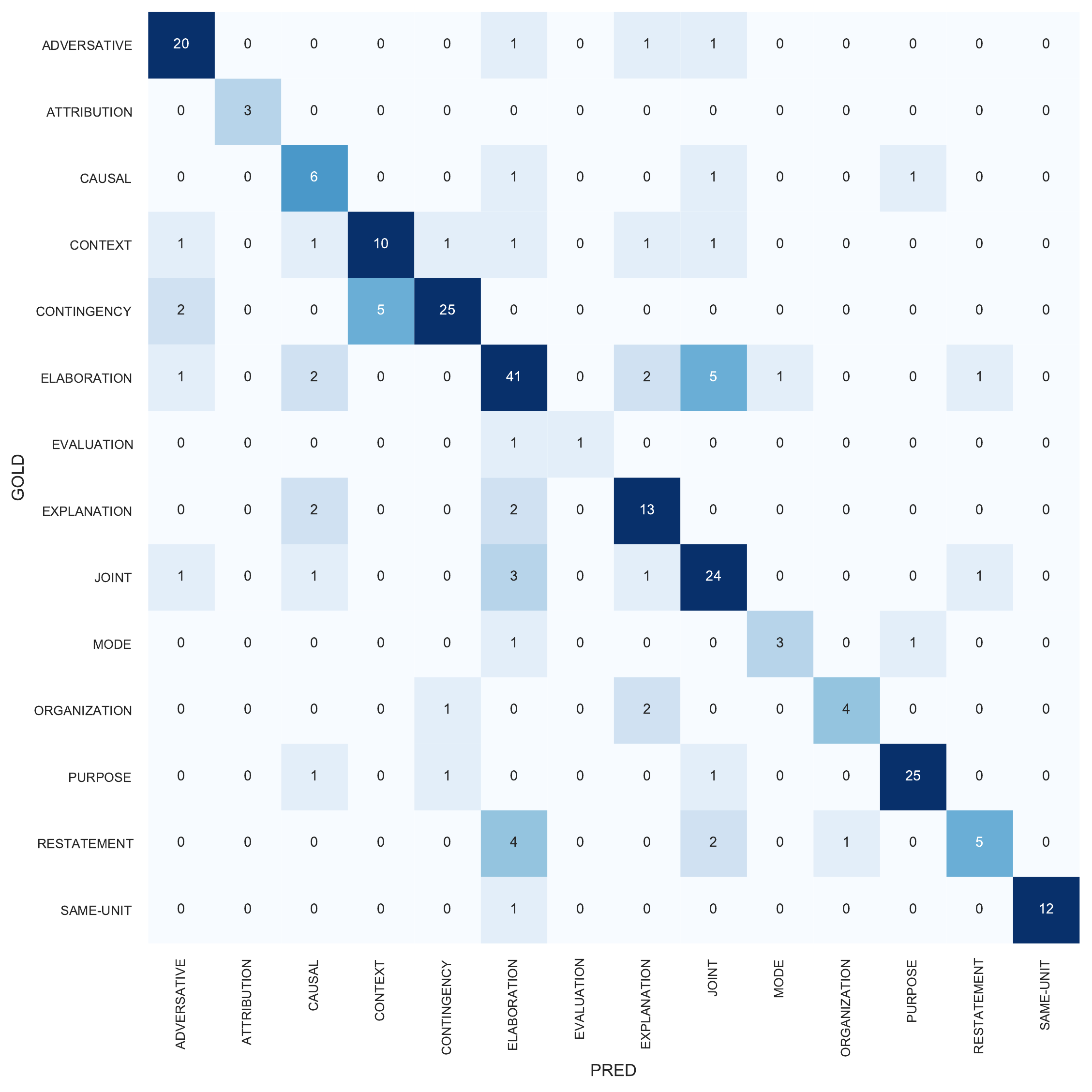}
    \vspace{-5pt}
    \caption{\textit{how-to} confusions (no attachment errors). }
    \vspace{-12pt}
    \label{fig:ood-whow-right-head}
\end{figure}

To further investigate genre-specific phenomena, we analyzed OOD parsing errors for \textit{how-to guides}, which had the worst OOD performance in Table \ref{tab:gum-ova-test-degradation}, using a model \textit{not} trained on this genre from \S\ref{subsec:GUM8-cross-genre-ova}. Figure \ref{fig:ood-whow-right-head} shows a confusion matrix produced by converting the automatic parses into the RST dependency representation following \citet{li-etal-2014-text}. We focus on relation instances with correct attachment predictions, although we also discuss cases involving attachment errors as well (Figure \ref{fig:ova-whow} in Appendix \ref{sec:appendix-ova-confmat}). We first observe in Figure \ref{fig:ood-whow-right-head} that there are five gold \textsc{Contingency} instances predicted to be \textsc{Context}: four out of five EDUs begin with a discourse marker (DM), `once' or `until', e.g.:

\vspace{-3pt}
\ex. [Don't use a joke] \SPSB{humans: \textsc{Contingency}}{parser:\quad\textsc{Context}} \quad\hfill\vspace{2mm}[\textbf{until} you're completely comfortable with it.] \label{ex:ood-whow-until}
\vspace{-4pt}

According to The Penn Discourse Treebank (PDTB3, \citealt{PrasadWebberLeeEtAl2019}), these prototypically signal spatio-temporal circumstances: 94\% of explicit instances with the DM `once' are annotated as \textsc{Temporal.Asynchronous} while less than 5\% of the instances (4/85) are annotated as \textsc{Contingency.Condition}. Similarly, 85\% of explicit instances with the DM `until' are annotated as \textsc{Temporal.Asynchronous} while 10\% are \textsc{Contingency.Condition}. However, \citet{liu-2019-beyond} found that these two DMs are frequently associated with \textsc{Contigency-Condition} in \textit{how-to guides} because they are essentially never narrative and always part of an instruction which is uncommon in \textit{news}. This suggests that although DMs are usually considered useful devices to identify certain relations, their usage differs across genres and is sometimes too ambiguous to form a reliable signal. We also observe that there are four gold \textsc{Restatement} confused with \textsc{Elaboration}: three of these go back to `definitional' \textsc{Restatements}, which are likely promoted by the genre's descriptive and explanatory properties. 

Figure \ref{fig:ova-whow} shows cases with attachment errors. One major error concerning global structure is seven gold \textsc{Organization} instances being confused with \textsc{Joint}. Two of the errors are unexpected given that there are strong graphical, structural, and semantic cues (see an example in Figure \ref{fig:ood-vs-gold-whow-examples-2023} in Appendix \ref{sec:appendix-ood-parser-outputs}). The remaining cases are harder to pin down: they are all imperatives, which are atypical for headings outside of \textit{how-to guides}, but normal and frequent in instructional texts (see an example in Figure \ref{fig:ood-vs-gold-whow-examples}). 
For complete by-genre confusion matrices (with attachment errors) of the multi-genre degradation experiments, see Appendix \ref{sec:appendix-ova-confmat}.

\section{Conclusion}
\label{sec:conclusion}

The analyses in this paper are meant to inform users of current parsers about what to expect from RST parsing in the wild. Through dozens of experimental runs we have shown a consistent picture: RST parsing has made impressive progress, but OOD degradation is still severe. Our results suggest prioritizing genre diversity in training data is crucial, not only to cover more text types as `in domain', but also to increase performance on unseen text types.  Rather than focusing only on the development of better models to beat the single genre RST-DT benchmark, robust RST parsing would be promoted most by creating more annotated data. 

We also hope this paper will motivate researchers to prioritize multi-genre benchmarks and OOD settings for RST parsing, and to explore algorithms, representations and features which better capitalize on joint training and foster generalizability, including using data from other sources and theories of discourse analysis (cf.~\citealt{braud-etal-2016-multi}).
We see great challenges for tackling errors which implicate complex pragmatic inference or document level reasoning, but we are also optimistic that as more data becomes available, we will be able to learn more and better representations of discourse structure.

\section*{Limitations}
\label{sec:limitations}

One limitation is that the scope of the error analysis presented here is limited, primarily due to space reasons. Although we included full relation confusion matrices (with attachment errors) in Appendix \ref{sec:appendix-ova-confmat}, the discussion is too limited to describe them in detail, and we encourage interested readers to explore and further compare gold and predicted trees (available in both~\texttt{.rs3} and~\texttt{.rsd} formats) on the multi-genre data, which we make available in the repository of this paper. 

Another limitation concerns the number of runs of the experiments: some scores were averaged over three rather than five runs. This is due to the fact that SpanBERT-base \cite{joshi-etal-2020-spanbert} is fairly large (110M parameters). Training each of our 20 models five times would consume very large amounts of GPU resources, which we feel is hard to justify both financially and environmentally. After completing this study we feel that three-run averaged scores should satisfy the need for reproducibility, though we did use five-run averages for establishing baseline scores and verifying particularly controversial results, such as the improvement on Nuclearity in the fine-tuning condition in \S\ref{subsec:joint} (i.e.~\textsc{sr-ft}). Previous work providing this information reported three-run averaged scores such as \citet{koto-etal-2021-top} while many papers did not include this information or mention whether the reported scores were averaged over multiple runs. 

\section*{Ethics Statement}
\label{sec:ethics-statements}
This work contributes to open source progress in RST discourse parsing, an area which has received less attention than some other NLP tasks, and which, at least in English, is currently suffering from a skewed focus towards the `standard' language of 1990-era Wall Street Journal writing. Previous work has shown that NLP systems retain strong lexical biases mirroring both period and author demographics \cite{shah-etal-2020-predictive}, meaning that if language technologies are not pushed to cover diverse data types robustly, they will inevitably perform more poorly on `non-standard' data, with possible discriminatory effects on under-represented populations ranging from the political (think opinion mining social media to guide policy) to financial (e.g.~higher/lower search hit rates for YouTube videos by small businesses). By promoting higher quality treatments of diverse language samples in this study covering Reddit, YouTube vlogs, and demographically diverse conversations, we hope to help level the playing field across text-types, demographics, and domains. 

We recognize that NLP research has a computing cost and carbon footprint, which motivates us to release the trained models in this work (preventing the need to retrain similar models), and to avoid extensive hyperparameter optimization which may not generalize to applications in the wild.
Specific model configurations such as hyperparameters and validation performance for the reported \texttt{test} results of RST-DT and GUM are detailed in Appendix \ref{sec:appendix-model-details}. 

Finally, we also recognize that NLP tools can be used to do harm. However, we expect that the type of analysis promoted here will do more good than harm by steering tool development away from adhering closely to outdated and narrow-domain data, which this work aims to broaden. Given that discourse parsers already exist, we view the push to reduce topical and authorial bias, as well as the public release of more resources, as net positives.


\bibliography{anthology,custom}
\bibliographystyle{acl_natbib}

\appendix

\section{Cross-Corpus Segmentation Scores}
\label{appendix:cross-corpus-seg-scores}

The scores in Table \ref{tab:cross-seg} (3 run average) show fairly modest degradation on cross-corpus EDU segmentation scores in both directions (a little worse for RST-DT$\rightarrow$GUM), using DisCoDisCo \cite{gessler-etal-2021-discodisco}, the winning EDU segmentation system from the 2021 DISRPT shared task \cite{zeldes-etal-2021-disrpt}.

\begin{table}[ht]
\centering
\resizebox{7cm}{!}{%
\begin{tabular}{ccccc}
\toprule
\textbf{train}  & \textbf{test} & \textbf{P} & \textbf{R} & \textbf{F1} \\
\midrule
RST-DT & RST-DT        & 95.16         & 94.64         & 94.90 \\
RST-DT & GUM V8        & 90.09         & 89.69         & 89.89 \\
GUM V8    & GUM V8     & 92.09         & 93.73         & 92.90 \\
GUM V8    & RST-DT     & 92.55         & 89.57         & 91.03 \\
\bottomrule
\end{tabular}
}%
\caption{Cross-Corpus Segmentation Performance. }
\label{tab:cross-seg}
\end{table}

\section{Breakdown of the GUM RST Data}
\label{sec:appendix:gum-content-breakdown}

The breakdown of genres in GUM V8 is shown in Table \ref{tab:gum8-data-overview}. Four of the genres in the corpus are still growing, and therefore include less material than other genres at present. 

\begin{table}[h!tb]
\centering
\resizebox{7cm}{!}{%
\begin{tabular}{lcllc}
\toprule
\textbf{genres} & \textbf{docs} & \textbf{tokens} & \textbf{EDUs} & \textbf{growing} \\
\midrule
\textit{academic} & 18 & 17,168 & 1,969 & \\
\textit{bio} & 20 & 18,209 & 2,066 & \\
\textit{fiction} & 19 & 17,508 & 2,458 & \\
\textit{how-to} & 19 & 17,085 & 2,367 & \\
\textit{interview} & 19 & 18,189 & 2,404 & \\
\textit{news} & 23 & 16,140 & 1,760 & \\
\textit{reddit} & 18 & 16,364 & 2,231 & \\
\textit{travel} & 18 & 16,513 & 1,785 & \\
\textit{conversation} & 9 & 10,451 & 1,878 & \checkmark \\
\textit{speech} & 10 & 10,827 & 1,249 & \checkmark \\
\textit{textbook} & 10 & 11,190 & 1,397 & \checkmark \\
\textit{vlog} & 10 & 11,200 & 1,543 & \checkmark \\
\hline
\textbf{total} & \textbf{193} & \textbf{180,844} & \textbf{23,107} \\
\bottomrule
\end{tabular}
}%
\caption{Overview of RST Data by Genre in GUM V8. }
\label{tab:gum8-data-overview}
\end{table}

\section{Model Configurations and Training Details}
\label{sec:appendix-model-details}


To ensure reproducibility and following the reproducibility criteria and checklist provided by EACL 2023,
we provide model configurations and training details relevant to the experimental results presented in this paper. 

For the \textsc{bottom-up} parser, the configurations of \texttt{SpanBERT-NoCoref} from \citet{guz-carenini-2020-coreference} was used as our base system, and we followed the hyperparameters and training settings therein across the board except the batch size due to memory limitations and embedding dimensions for organizational features.\footnote{Based on our reproducible results of this model (see Table \ref{tab:overview-rst-parsers}) and comparing them to the reported results in the original paper, we believe that the change of these two configurations do not impact model performance significantly.} Specifically, AdamW was used as the optimizer with a learning rate of 1\textit{e}\textsuperscript{--5} for SpanBERT-base \cite{joshi-etal-2020-spanbert} and 2\textit{e}\textsuperscript{--4} for model-specific components \citep{guz-carenini-2020-coreference}. Batch size was set to 1 (as opposed to 5 in the original implementation) and there were 20 epochs for each run. The organizational features used in \citet{guz-carenini-2020-coreference} followed \citet{wang-etal-2017-two} and were represented as binary features in a learnable 5-dimensional embedding (as opposed to 10 in the original implementation). A vector of zeros of the same shape was used when a given feature is unavailable for the aforementioned organizational features as well as the categorical features experimented in \S\ref{subsec:joint}, which all had an embedding dimension of 10. 

For the \textsc{top-down} parser, overall we followed the original hyperparameters and training settings therein; however, due to memory limitations, we modified the batch size to 3 from 12.
The same \texttt{dev} set of respective corpora was used during training as in the training of the \textsc{bottom-up} parser from \citet{guz-carenini-2020-coreference}.
Additionally, XLM-RoBERTa-base \cite{conneau-etal-2020-unsupervised} was used as the language backbone. 

All the training sessions were conducted using 1 NVIDIA Tesla T4 GPU on Google Cloud Platform. Table \ref{tab:corresponding-validation-scores} shows the corresponding validation performance
on RST-DT \texttt{test} and GUM V8 \texttt{test} reported in Tables \ref{tab:cross-corpus-guz} and \ref{tab:rstdt-categorical} respectively using the \textsc{bottom-up} parser. Table \ref{tab:corresponding-validation-scores-dmrst} presents the corresponding validation performance (averaged over 5 runs) on RST-DT \texttt{test} and GUM V8 \texttt{test} reported in Table \ref{tab:cross-corpus-dmrst}.

\begin{table}[htp]
\centering
\resizebox{\columnwidth}{!}{%
\begin{tabular}{l|cccc}
\toprule
& \textbf{S} & \textbf{N} & \textbf{R} & \textbf{\# runs} \\ \midrule
RST-DT \textsc{baseline}  & 76.0  & 64.9  & 55.2  & 5  \\
GUM \texttt{test}  & 67.0  & 54.5  & 45.9  & 5 \\ \hline 
RST-DT \textsc{concat}    & 76.4  & 65.7  & 56.0  & 3  \\
RST-DT \textsc{flair-label}   & 76.3       & 65.1       & 54.9       & 3  \\
RST-DT \textsc{sr-label}      & 76.3       & 64.7       & 55.2       & 3  \\
RST-DT \textsc{sr-graph} & 76.6       & 65.5       & 55.5       & 3       \\
RS-TDT \textsc{sr-ft}    & 76.3       & 64.9       & 55.4       & 5       \\
\bottomrule
\end{tabular}%
}
\caption{Validation Performance on RST-DT \texttt{test} and GUM \texttt{test} Results in Tables \ref{tab:cross-corpus-guz} and \ref{tab:rstdt-categorical} (i.e.~\textsc{bottom-up}).}
\label{tab:corresponding-validation-scores}
\end{table}

\begin{table}[htp]
\centering
\resizebox{\columnwidth}{!}{%
\begin{tabular}{l|cccc}
\toprule
& \textbf{S} & \textbf{N} & \textbf{R} & \textbf{\# runs} \\ \midrule
RST-DT \textsc{baseline} & 75.3       & 65.0       & 55.9       & 5  \\
GUM \texttt{test}        & 71.3       & 58.6       & 50.1       & 5  \\ \bottomrule
\end{tabular}%
}
\caption{Validation Performance on RST-DT \texttt{test} and GUM \texttt{test} Results in Table \ref{tab:cross-corpus-dmrst} (i.e.~\textsc{top-down}).}
\label{tab:corresponding-validation-scores-dmrst}
\end{table}

Note that since RST-DT does not have an established \texttt{dev} set, 10\% of training data stratified by the number of EDUs in each document is used as the \texttt{dev} set following \citet{guz-carenini-2020-coreference} and is held constant across all the RST-DT-related experiments in this work. The list of the document names used in the \texttt{dev} set is provided in the code repository.\footnote{\url{https://github.com/janetlauyeung/crossGENRE4RST}} Additionally, since the \texttt{test} documents used in all the experiments in \S\ref{subsec:GUM8-cross-genre-ova} and \S\ref{subsec:genre-variety-fixed-sized} are OOD data, the corresponding validation performance is not applicable and thus not reported. Information on the average runtime of each epoch for each model is provided in Table \ref{tab:avg-run-time} below.

\begin{table}[ht]
\centering
\resizebox{\columnwidth}{!}{%
\begin{tabular}{llll}
\toprule
\multicolumn{4}{c}{\multirow{2}{*}{\textbf{RST-DT}}}    \\
\multicolumn{4}{c}{}                                    \\ \midrule
\textbf{model} & \multicolumn{1}{l|}{\textbf{avg. runtime}}    & \textbf{model} & \textbf{avg. runtime}    \\ \midrule
\textsc{baseline}       & \multicolumn{1}{l|}{2 hours 53 mins} & \textsc{sr-label}       & 2 hours 53 mins \\
\textsc{concat}         & \multicolumn{1}{l|}{5 hours 52 mins} & \textsc{sr-graph}       & 2 hours 50 mins \\
\textsc{flair-label}    & \multicolumn{1}{l|}{2 hours 55 mins} & \textsc{sr-ft}          & 2 hours 56 mins \\ \midrule
\multicolumn{4}{c}{\multirow{2}{*}{\textbf{GUM}}}                \\
\multicolumn{4}{c}{}                                    \\ \midrule
\multicolumn{1}{l}{\textbf{model}} & \multicolumn{1}{l|}{\textbf{avg. runtime}} & \multicolumn{1}{l}{\textbf{model}} & \multicolumn{1}{l}{\textbf{avg. runtime}} \\ \midrule
GUM \texttt{test}       & \multicolumn{1}{l|}{3 hours}         & No Reddit      & 2 hours 43 mins \\
\textsc{all-large}        & \multicolumn{1}{l|}{2 hours 15 mins} & No Voyage      & 2 hours 47 mins \\
No Academic    & \multicolumn{1}{l|}{2 hours 43 mins} & No How-to        & 2 hours 45 mins \\
No Bio         & \multicolumn{1}{l|}{2 hours 45 mins} & C1       & 56 mins         \\
No Fiction     & \multicolumn{1}{l|}{2 hours 44 mins} & C2       & 56 mins         \\
No Interview   & \multicolumn{1}{l|}{2 hours 40 mins} & C3       & 54 mins         \\
No News        & \multicolumn{1}{l|}{2 hours 47 mins} &                &                 \\ \bottomrule
\end{tabular}%
}
\caption{Average Runtime of Every Training Session's Epoch for Each Model using the \textsc{bottom-up} Parser. } 
\label{tab:avg-run-time}
\vspace{-10pt}
\end{table}

\section{Relation Mapping}
\label{sec:appendix-relation-mapping}

In the interest of reproducibility, Table \ref{tab:gum8-rstdt-mapping} gives the exact relation mapping used for cross-corpus experiments in which relation labels were targeted.

\begin{table}[ht]
\centering
\resizebox{\columnwidth}{!}{%
\begin{tabular}{lll}
\toprule
\textbf{\begin{tabular}[c]{@{}l@{}}GUM V8 \\ Relations\end{tabular}} &
\textbf{\begin{tabular}[c]{@{}l@{}}GUM V8 \\ Classes\end{tabular}} &
\textbf{\begin{tabular}[c]{@{}l@{}}Corresponding \\ RST-DT Classes \end{tabular}} \\ \midrule 
adversative-antithesis   & Adversative  & Contrast             \\
adversative-concession   & Adversative  & Contrast             \\
adversative-contrast     & Adversative  & Contrast             \\
attribution-positive     & Attribution  & Attribution          \\
attribution-negative     & Attribution  & Attribution          \\
causal-cause             & Causal       & Cause                \\
causal-result            & Causal       & Cause                \\
context-background       & Context      & Background           \\
context-circumstance     & Context      & Background           \\
contingency-condition    & Contingency  & Condition            \\
elaboration-attribute    & Elaboration  & Elaboration          \\
elaboration-additional   & Elaboration  & Elaboration          \\
explanation-evidence     & Explanation  & Explanation          \\
explanation-justify      & Explanation  & Explanation          \\
explanation-motivation   & Explanation  & Explanation          \\
evaluation-comment       & Evaluation   & Evaluation           \\
joint-disjunction        & Joint        & Joint                \\
joint-list               & Joint        & Joint                \\
joint-sequence           & Joint        & Temporal             \\
joint-other              & Joint        & Topic-Change         \\
mode-manner              & Mode         & Manner-Means         \\
mode-means               & Mode         & Manner-Means         \\
organization-heading     & Organization & Textual-Organization \\
organization-phatic      & Organization & Topic-Comment        \\
organization-preparation & Organization & Textual-Organization \\
purpose-attribute        & Purpose      & Elaboration          \\
purpose-goal             & Purpose      & Enablement           \\
restatement-partial      & Restatement  & Summary              \\
restatement-repetition   & Restatement  & Summary              \\
topic-question           & Topic        & Topic-Comment        \\
topic-solutionhood       & Topic        & Topic-Comment        \\
same-unit                & same-unit    & Same-Unit            \\ \bottomrule
\end{tabular}%
}
\caption{Relation Mapping of GUM V8 to RST-DT. }
\vspace{-10pt}
\label{tab:gum8-rstdt-mapping}
\end{table}

\section{Data Description for the GUM OOD Multi-Genre Experiments}
\label{sec:apendix-data-overview-ova}

Table \ref{tab:ova-data-description} details the number of genres, documents, and EDUs used for training the models in the ten experiments in \S\ref{subsec:GUM8-cross-genre-ova}. For maximum reliability, testing was always conducted on the held-out genres' standard \texttt{dev}+\texttt{test} partitions specified in the official GUM V8 release (which is also provided in the repository of this paper), since the \texttt{dev} set for the targeted genres cannot be used for early stopping to simulate real OOD data in the wild. 

\begin{table}[ht]
\centering
\resizebox{6cm}{!}{%
\begin{tabular}{l|ccc}
\toprule
\textbf{models} & \textbf{genres} & \textbf{docs} & \textbf{EDUs} \\ \midrule
No Academic  & 11 & 131 & 16,088 \\
No Bio       & 11 & 129 & 15,901 \\
No Fiction   & 11 & 130 & 15,640 \\
No Interview & 11  & 130 & 15,599 \\
No News      & 11 & 126 & 16,252 \\
No Reddit    & 11 & 131 & 15,892 \\
No Voyage    & 11 & 131 & 16,133 \\
No How-to    & 11 & 130 & 15,672 \\
\textsc{all-large} & 8 & 122 & 13,703 \\
GUM \texttt{test}  & 12 & 145 & 17,610 \\ \bottomrule
\end{tabular}%
}
\caption{Training Data Composition used in \S\ref{subsec:GUM8-cross-genre-ova}. } 
\label{tab:ova-data-description}
\end{table}

\section{An Example OOD Parser Output}
\label{sec:appendix-ood-parser-outputs}

\begin{figure}[ht]
    \centering
    \subfloat[Parser Output Fragment from the \textit{No How-to} Model.]{\includegraphics[width=\columnwidth]{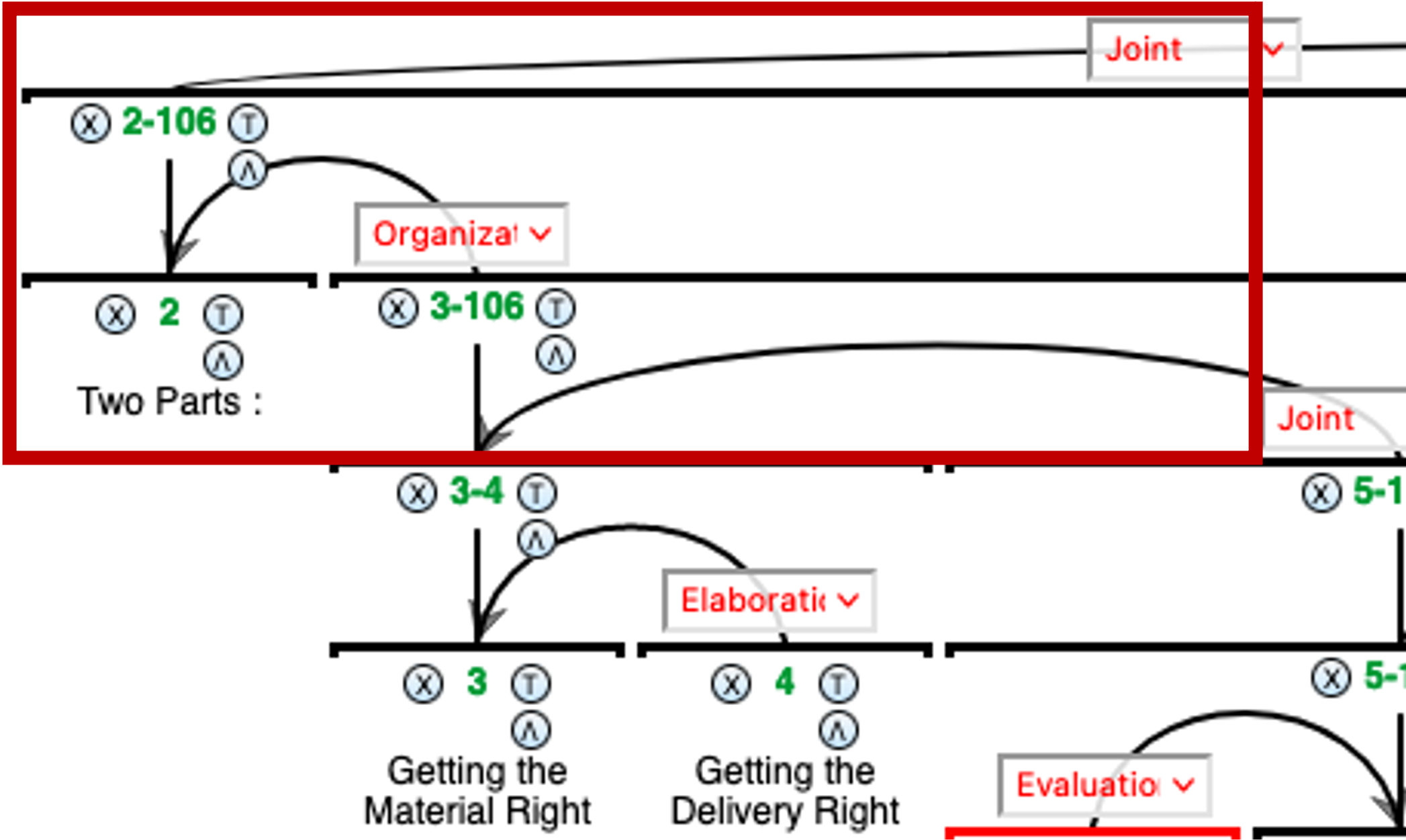}}\label{fig:ood-whow-2023-pred}
    \quad
    \subfloat[The Corresponding Gold RST Annotation. ]{\includegraphics[width=\columnwidth]{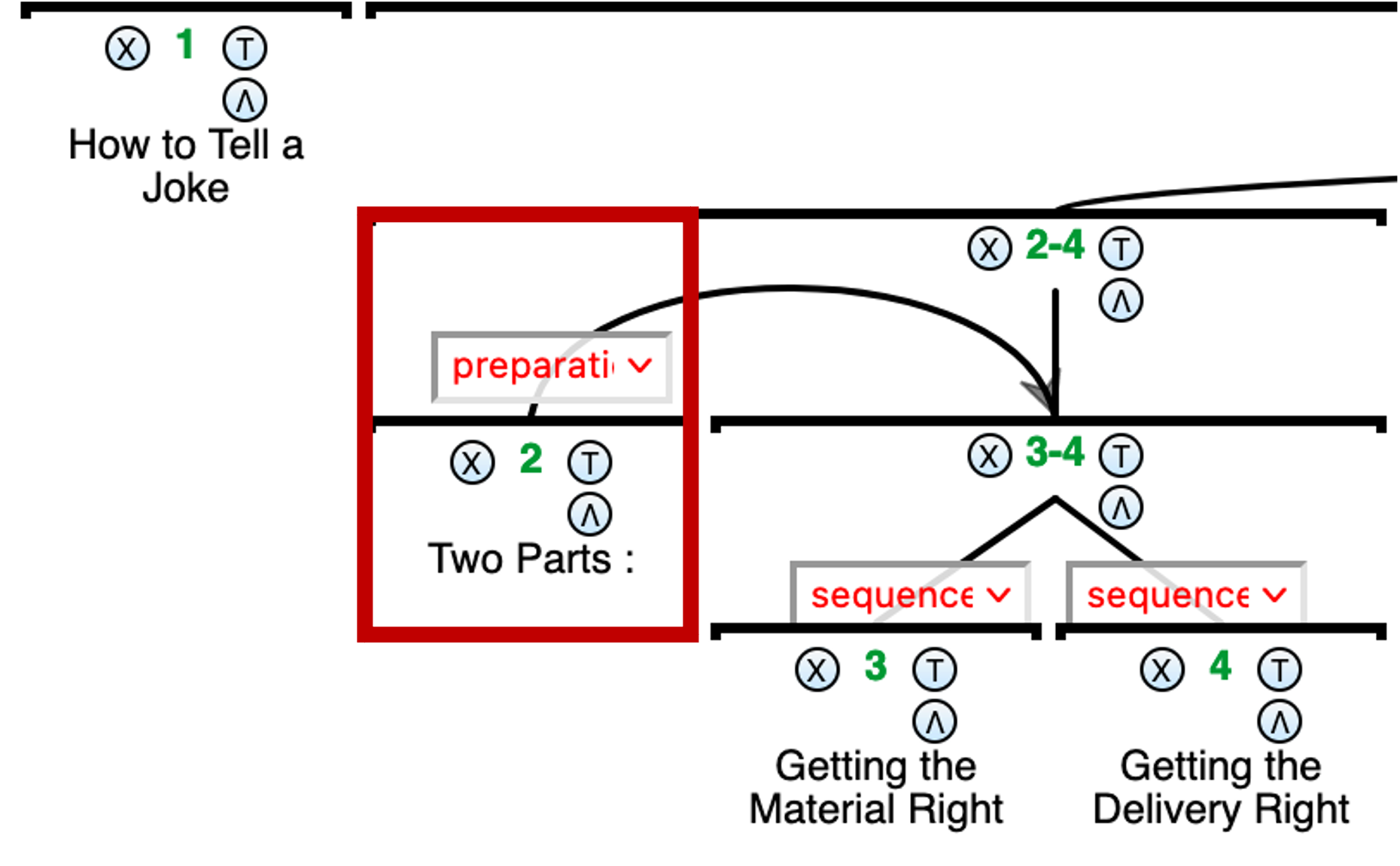}}\label{fig:ood-whow-2023-gold}
    
    \caption{An Example of an OOD Parser Output vs.~the Corresponding Gold Annotations from \textit{how-to guides}: \textsc{Organization} vs.~\textsc{Joint}. }
    \label{fig:ood-vs-gold-whow-examples-2023}
\end{figure}

\begin{figure*}[htp]
    \centering
    \subfloat[Parser Output Fragment from the \textsc{ova} (No How-to) Model in \S\ref{subsec:GUM8-cross-genre-ova}.]{\includegraphics[width=\textwidth,scale=1]{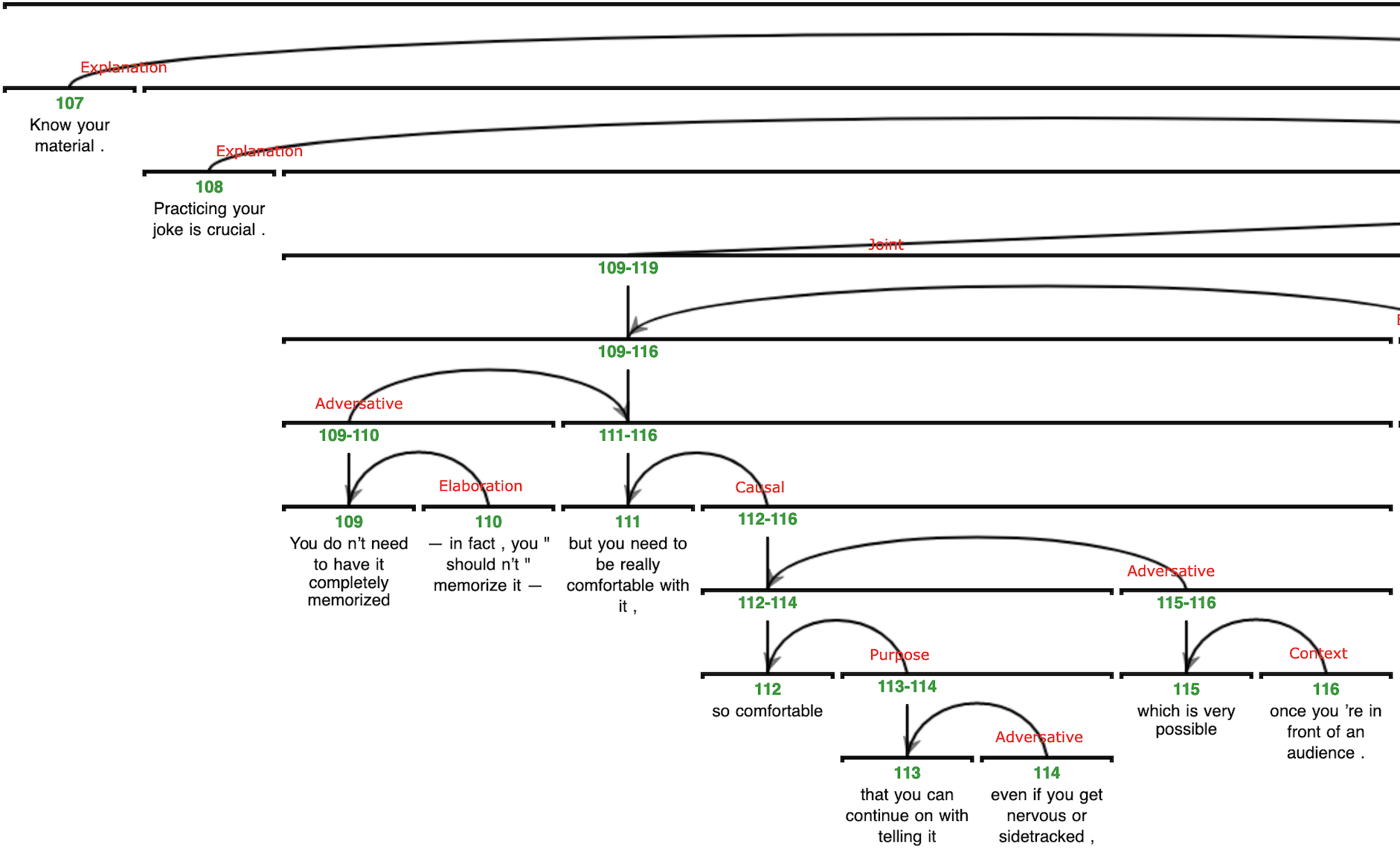}}\label{fig:ood-whow}
    \quad\quad
    \subfloat[The Corresponding Gold RST Annotation. ]{\includegraphics[width=\textwidth,scale=1]{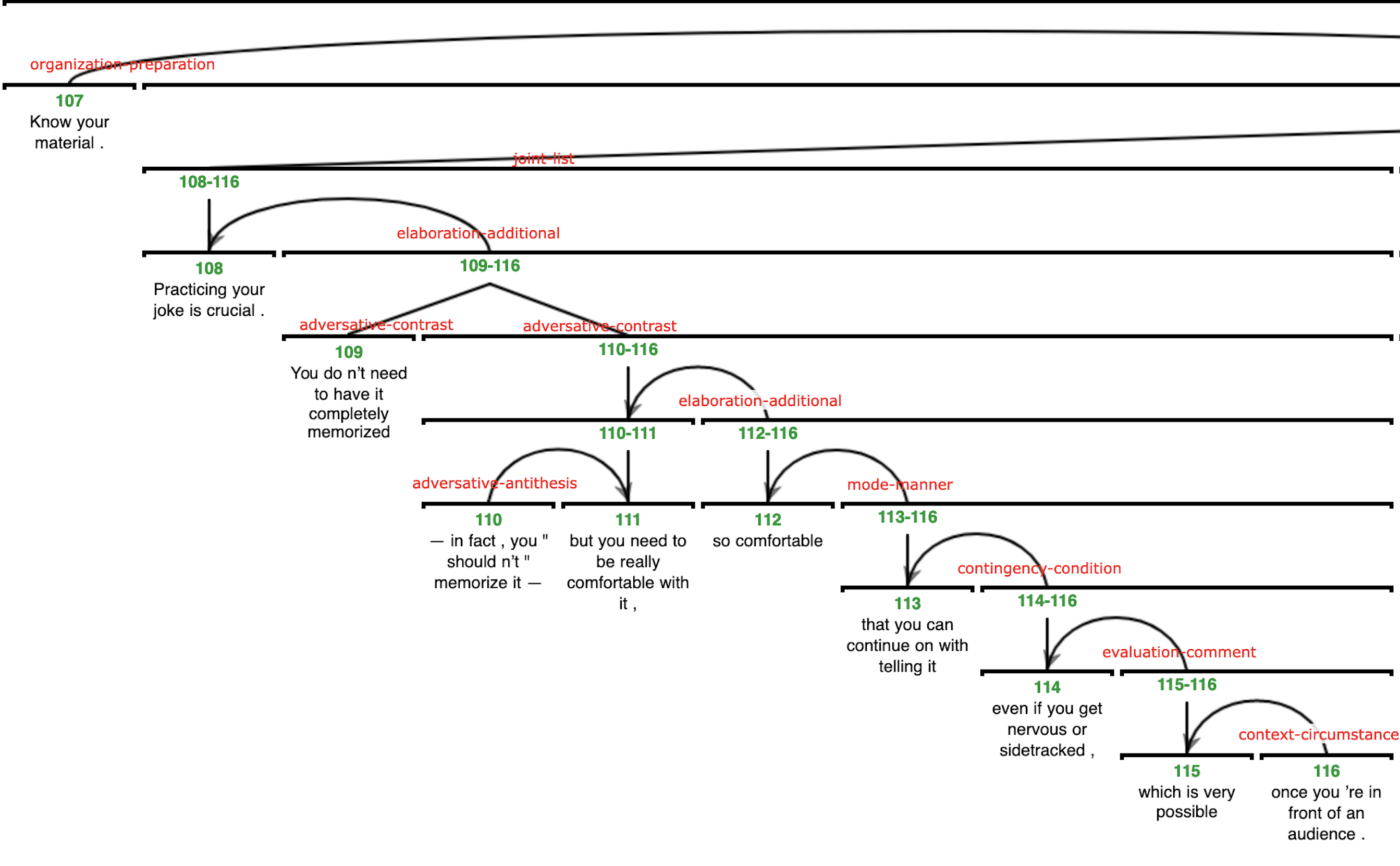}}\label{fig:gold-whow}
    
    \caption{An Example of an OOD Parser Output vs.~the Corresponding Gold Annotations from \textit{how-to guides}. This fragment is selected from the document GUM\_whow\_joke, which gives advice on how to tell jokes. }
    \label{fig:ood-vs-gold-whow-examples}
    
\end{figure*}

Figure \ref{fig:ood-vs-gold-whow-examples-2023} provides a comparison of a fragment of parser output versus the corresponding gold annotation to exemplify the distinction between gold \textsc{Organization} and predicted \textsc{Joint} discussed in Section \ref{sec:error-analysis}. Additionally, Figure \ref{fig:ood-vs-gold-whow-examples} provides a fragment of parser output from one of the \textsc{ova} models, No How-to, which exemplifies some of the observations and discussions brought up in the error analysis such as the most erroneous relation class \textsc{Organization} in how-to guides (EDU 107). Moreover, these figures also demonstrate that although discourse markers (DMs) are cues in many cases, they can lead to errors if they are ambiguous, by distracting the parser from other (non-DM) signals such as the misidentified \textsc{Elaboration} of EDU 110, where the gold \textsc{Organization} is signaled by the colon and the numerical matching of `Two parts' to the two nucleaus EDUs, 3 and 4 in Figure \ref{fig:ood-vs-gold-whow-examples-2023}; and the \textsc{Causal} of EDUs 112-116 in Figure \ref{fig:ood-vs-gold-whow-examples}, where an intensifying `so' in 112 (`so comfortable') may have misled the parser into a resultative `so' reading. 

\section{Confusion Matrices}
\label{sec:appendix-ova-confmat}

Figures \ref{fig:ova-academic}--\ref{fig:ova-whow} show confusion matrices for all the non-growing genres from their corresponding \textsc{ova} models, and Figures \ref{fig:ova-conversation}--\ref{fig:ova-vlog} are for all the growing genres based on the automatic parses from the \textsc{all-large} model where the training data contains only the 8 non-growing genres. Note that all the matrices were produced by converting the automatic parses into the RST dependency representation following \citet{li-etal-2014-text} and do not reflect attachment errors. The conversion code is available at \url{https://github.com/amir-zeldes/rst2dep}.

\begin{figure*}[htp]
  \centering
  \subfloat[\textit{academic}]{\includegraphics[width=70mm,scale=1]{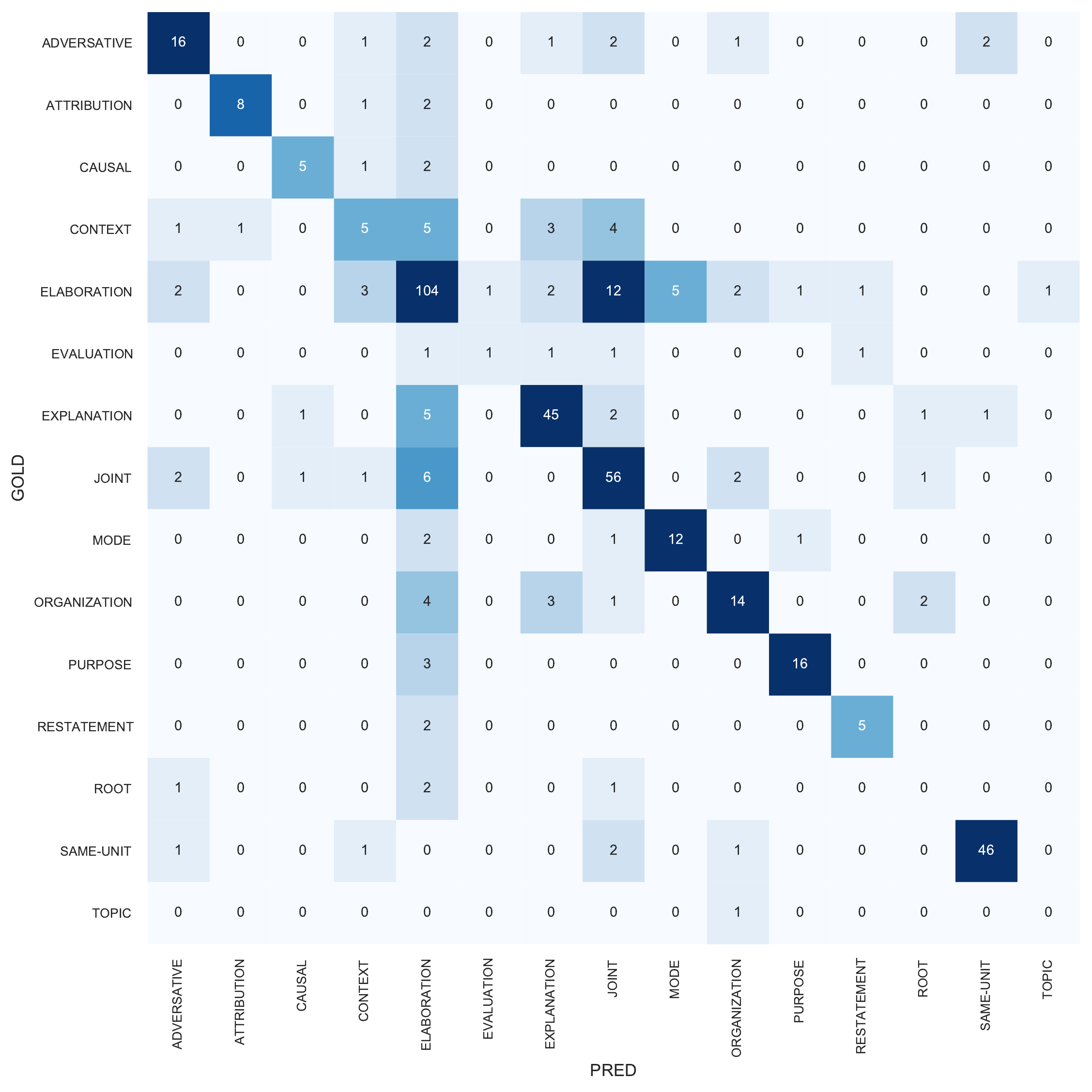}\label{fig:ova-academic}}
  \hfill 
  \subfloat[\textit{bio}]{\includegraphics[width=70mm,scale=1]{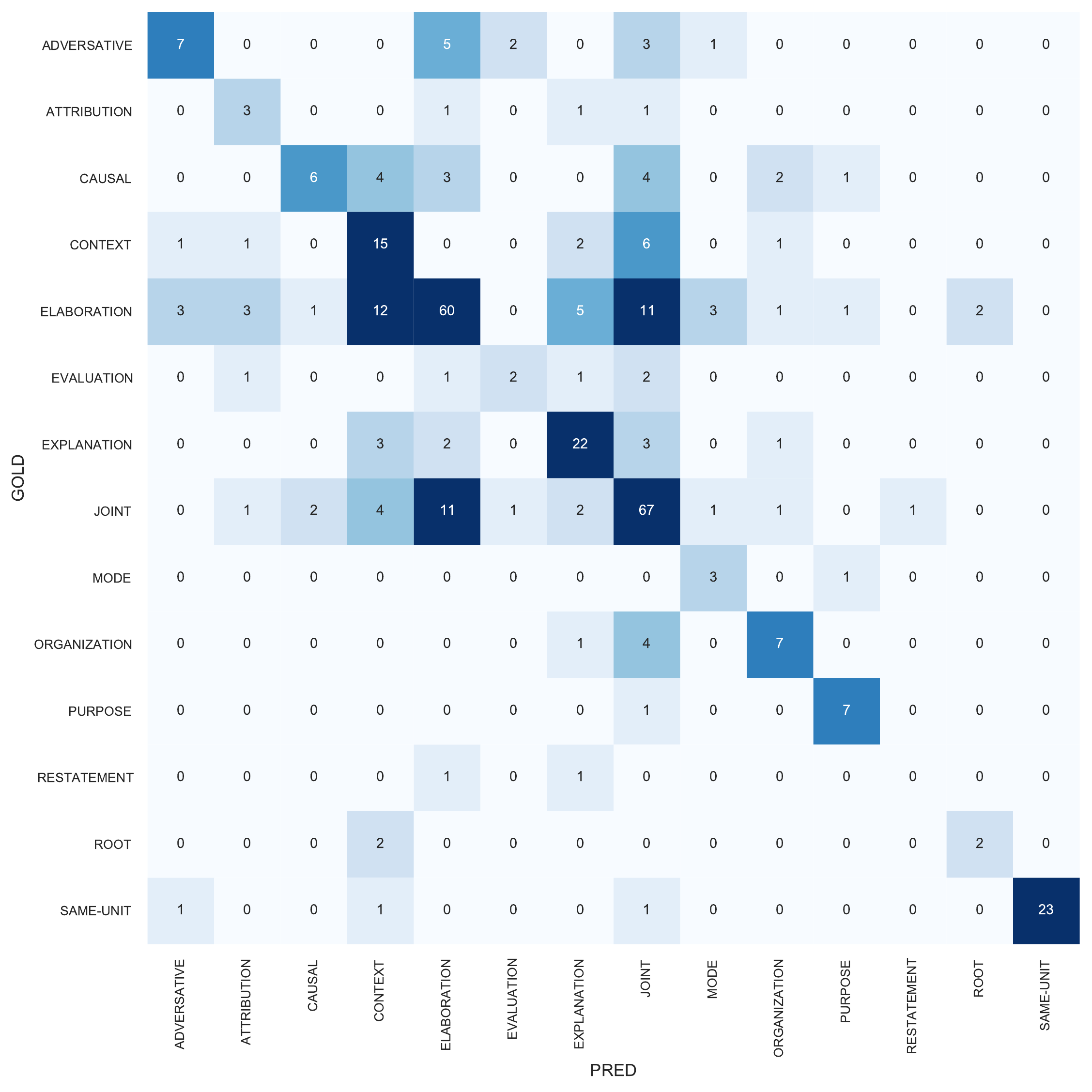}\label{fig:ova-bio}}
  \hfill 
  \subfloat[\textit{fiction}]{\includegraphics[width=70mm,scale=1]{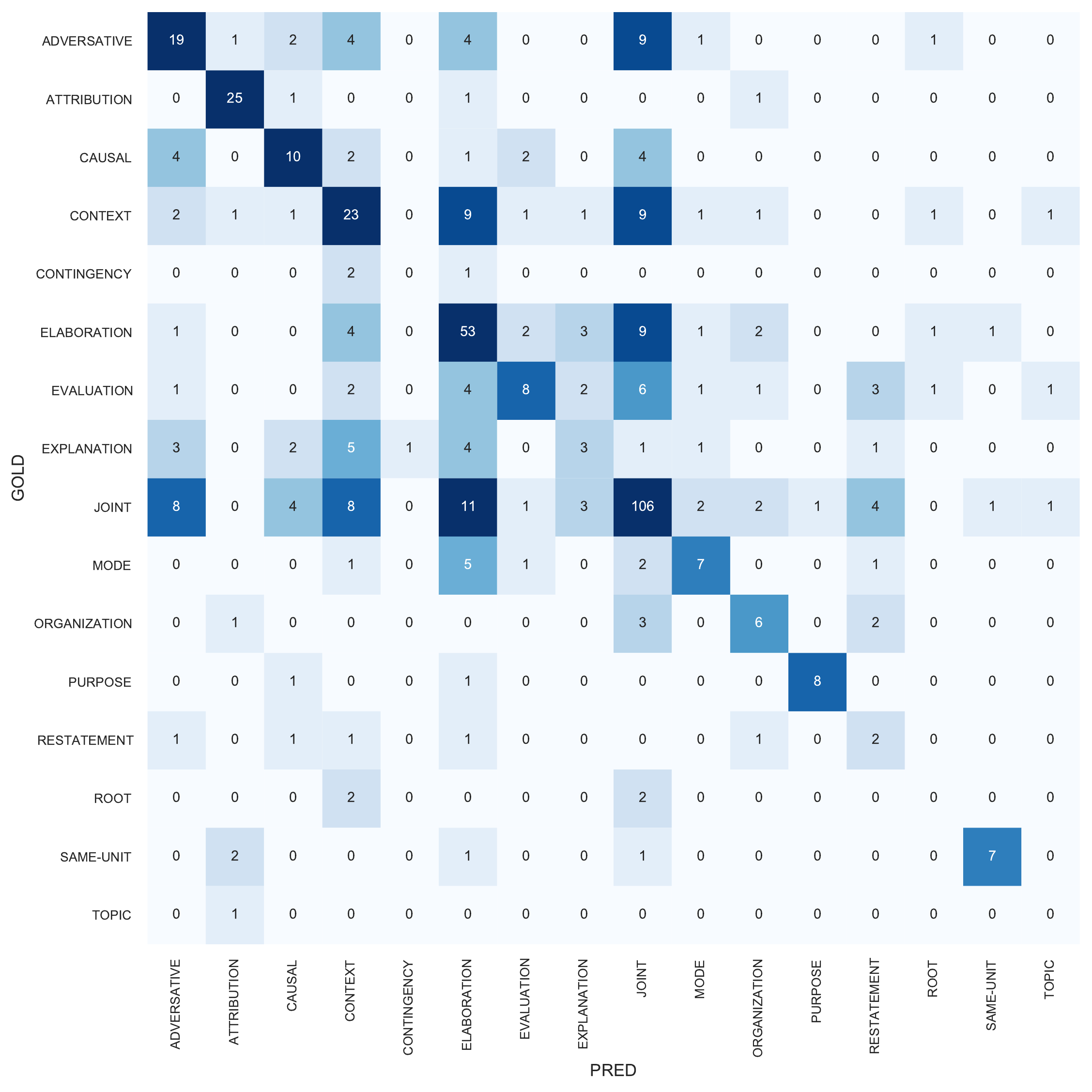}\label{fig:ova-fiction}}
  \hfill 
  \subfloat[\textit{interview}]{\includegraphics[width=70mm,scale=1]{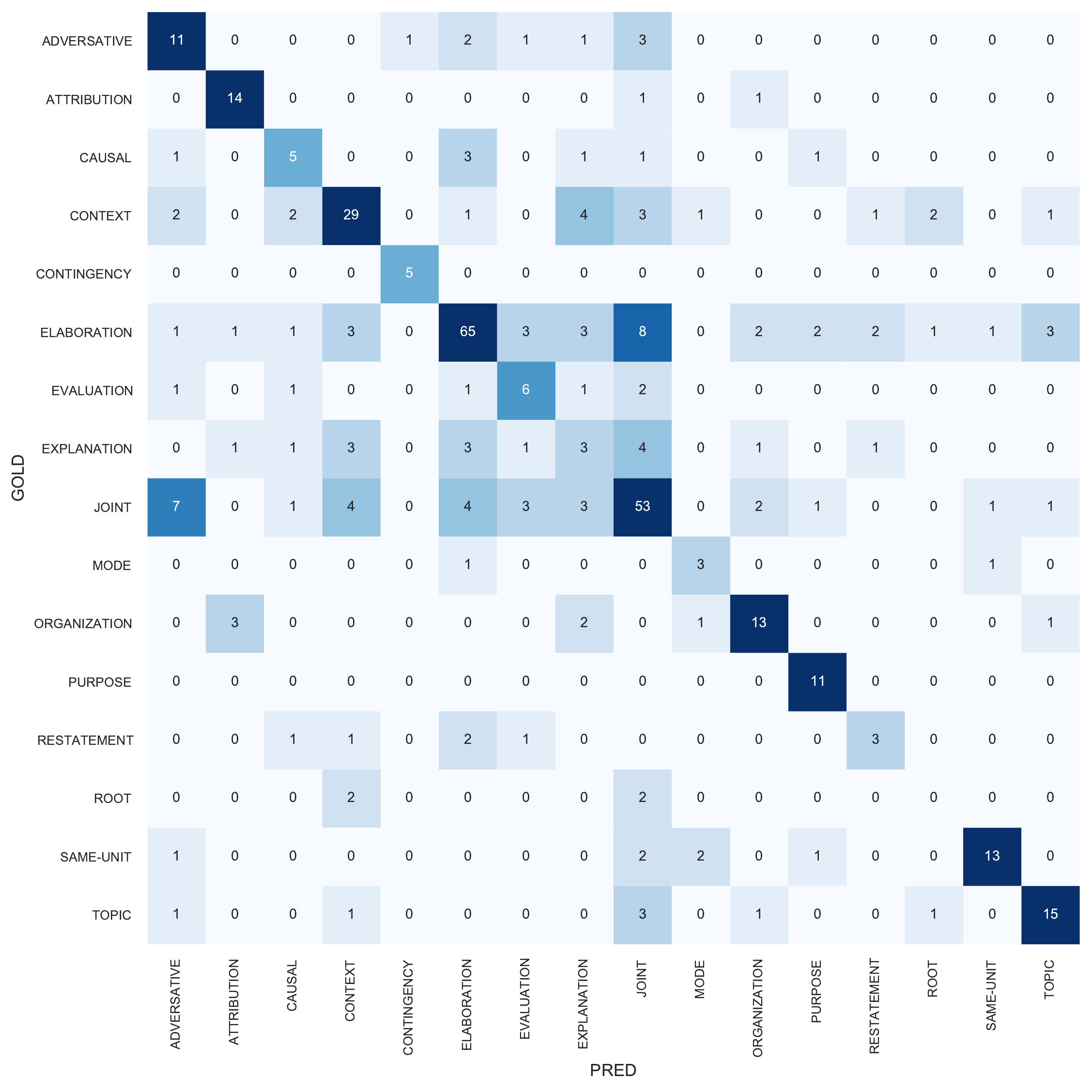}\label{fig:ova-interview}}
  \hfill
  \subfloat[\textit{news}]{\includegraphics[width=70mm,scale=1]{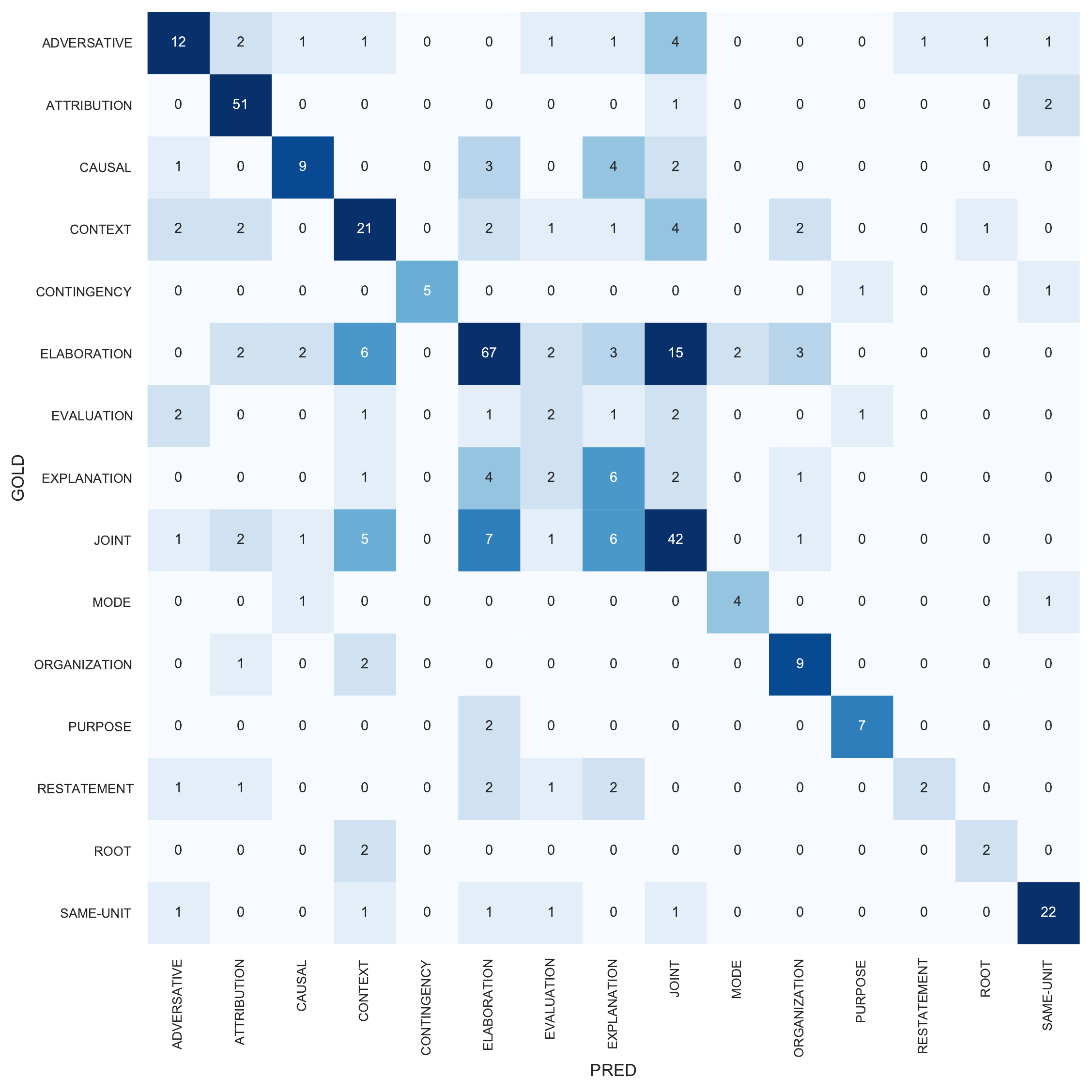}\label{fig:ova-news}}
  \hfill 
  \subfloat[\textit{reddit}]{\includegraphics[width=70mm,scale=1]{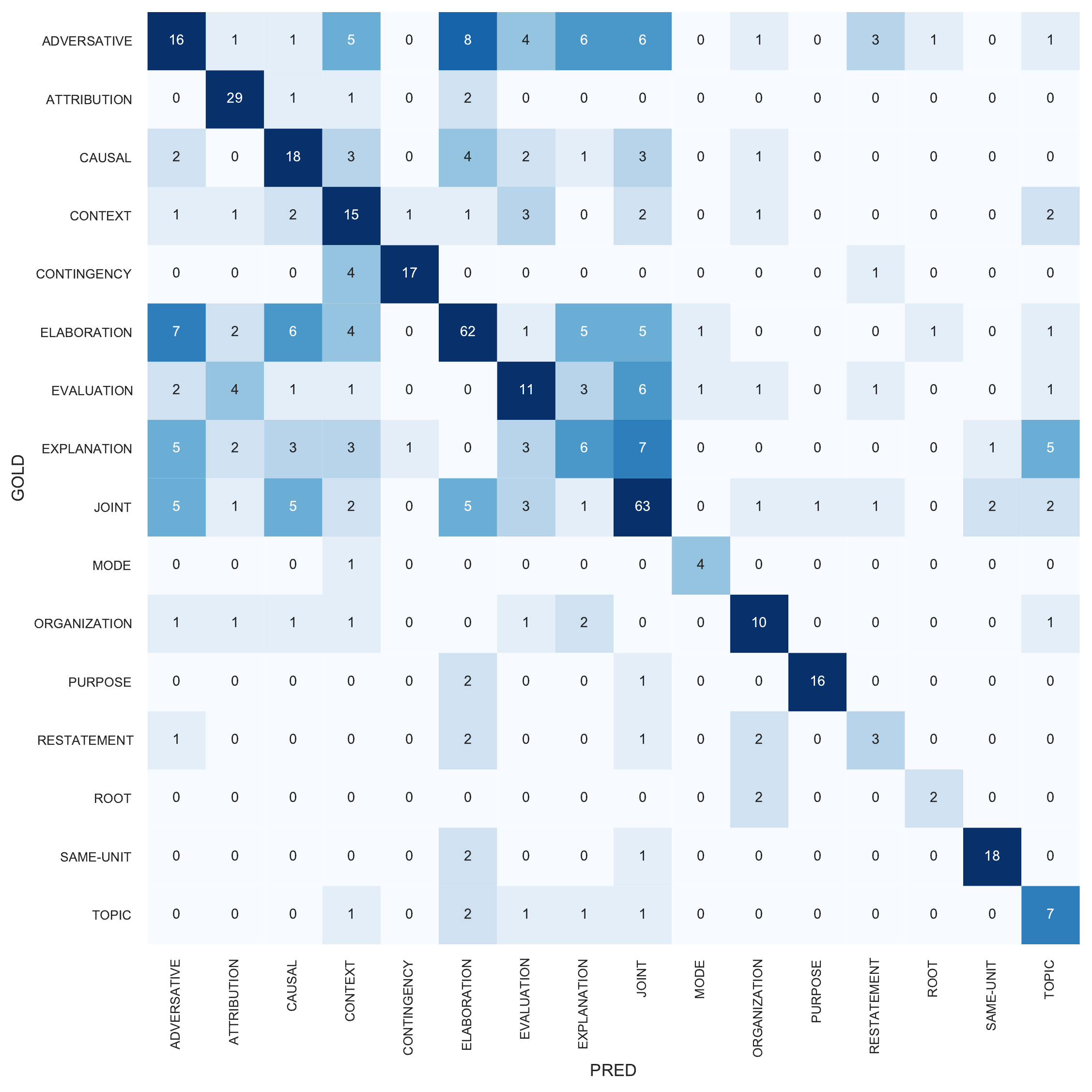}\label{fig:ova-reddit}}
  
  \label{fig:confusion-matrices-ova-experiment}
  \caption{Confusion Matrices for All Genres from their \textsc{ova} Models or the \textsc{all-large} Model from \S\ref{subsec:GUM8-cross-genre-ova}.}
\end{figure*}

\begin{figure*}[htp]\ContinuedFloat
  \centering
  \subfloat[\textit{voyage}]{\includegraphics[width=70mm,scale=1]{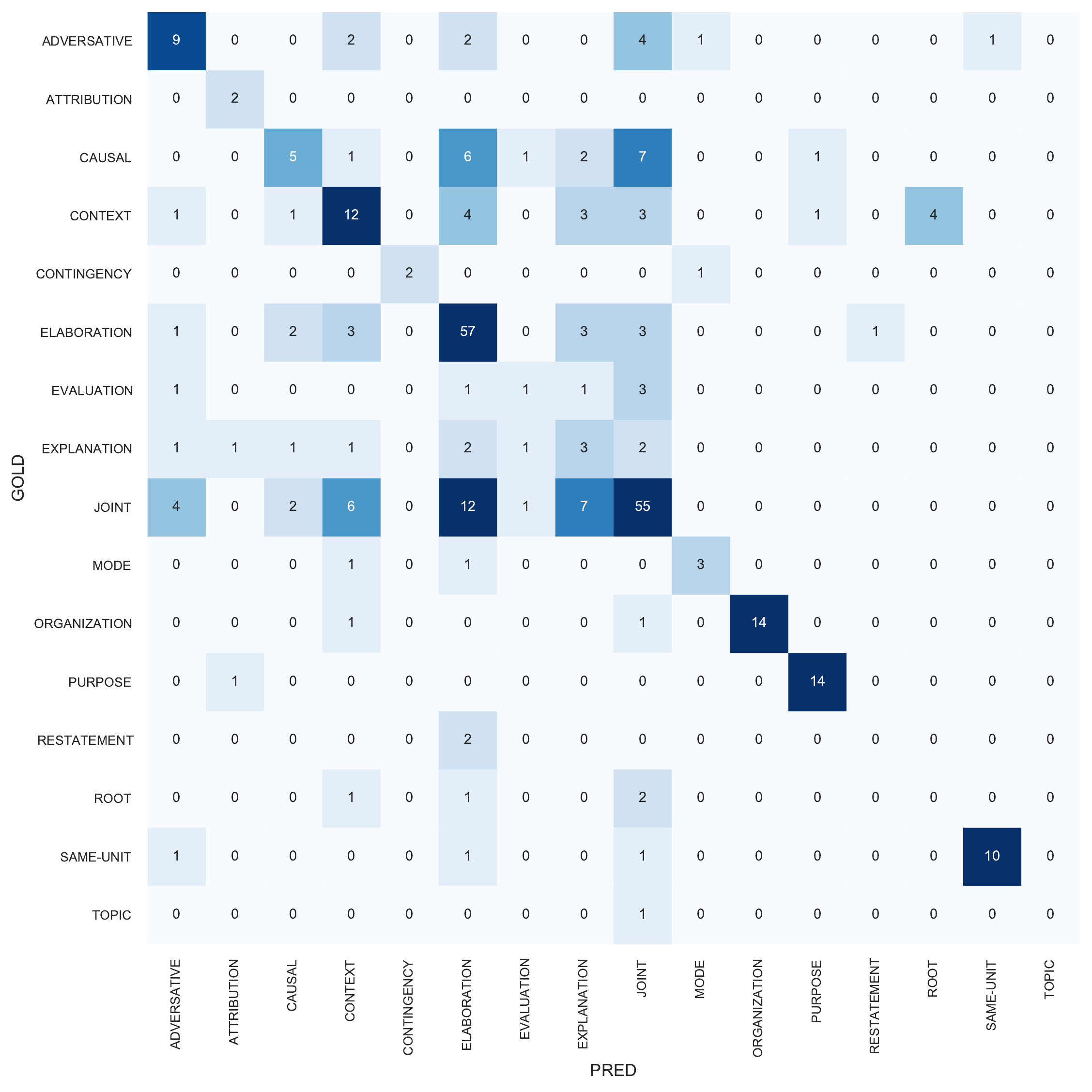}\label{fig:ova-voyage}}
  \hfill 
  \subfloat[\textit{whow}]{\includegraphics[width=70mm,scale=1]{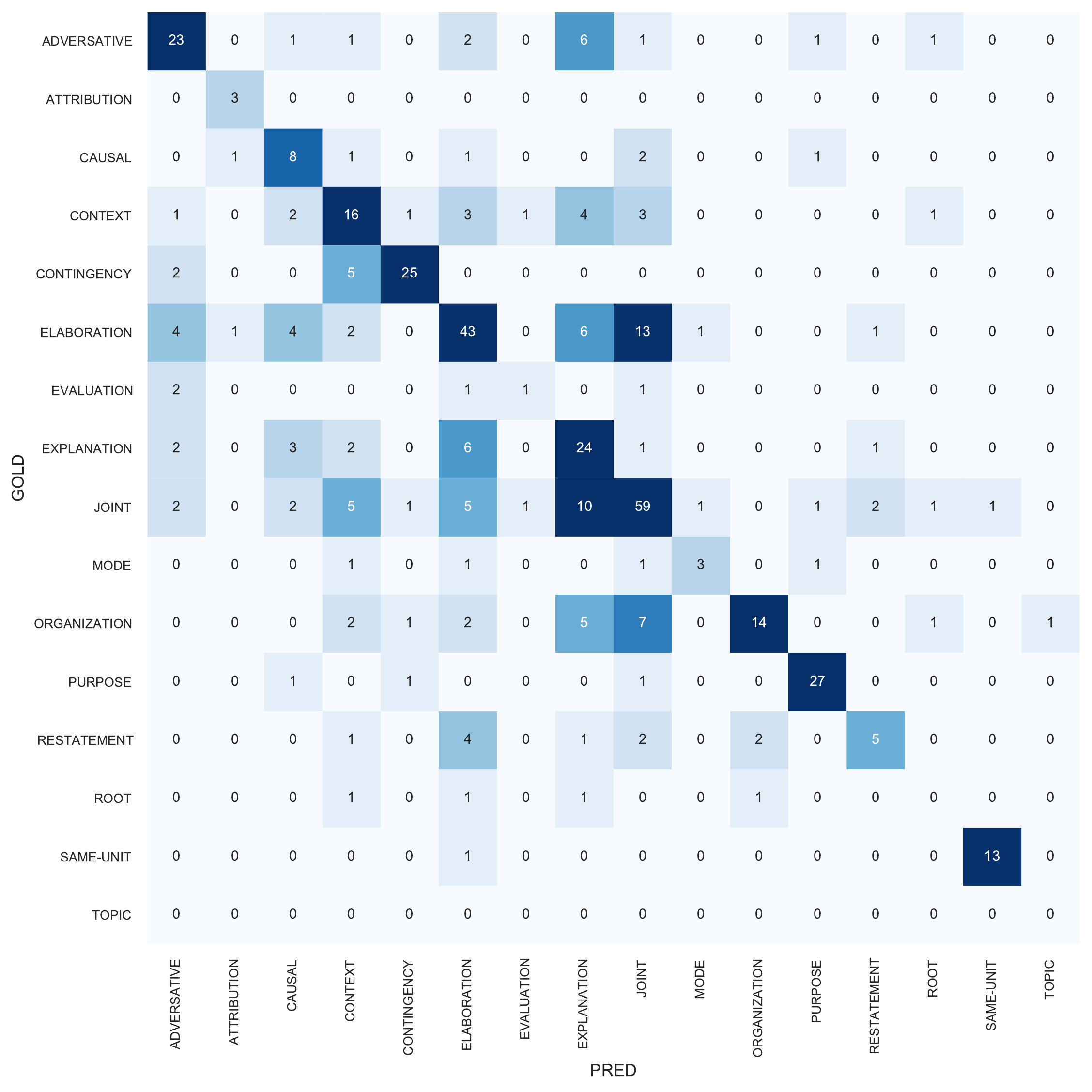}\label{fig:ova-whow}}
  \hfill
  \subfloat[\textit{conversation}]{\includegraphics[width=70mm,scale=1]{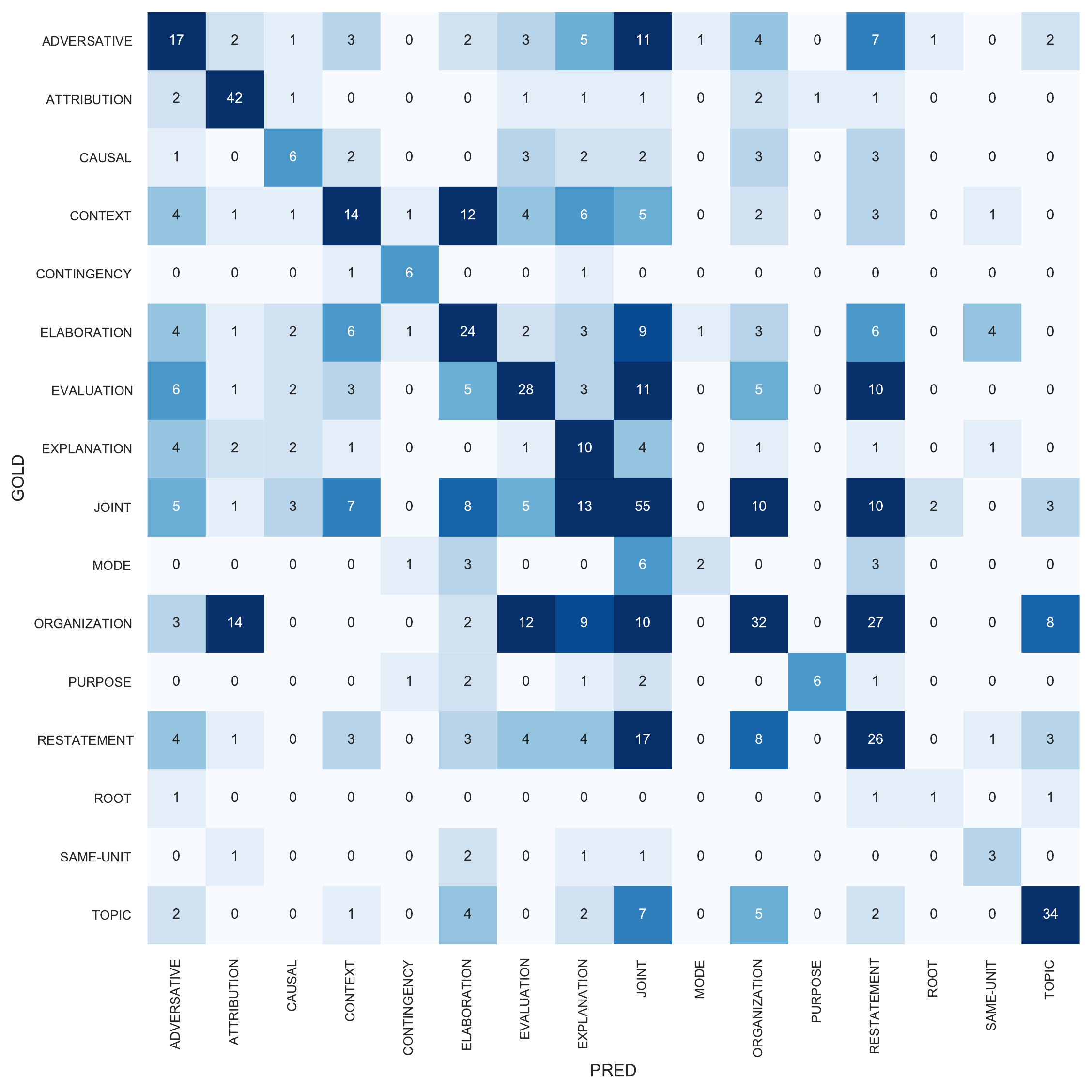}\label{fig:ova-conversation}}
  \hfill 
  \subfloat[\textit{speech}]{\includegraphics[width=70mm,scale=1]{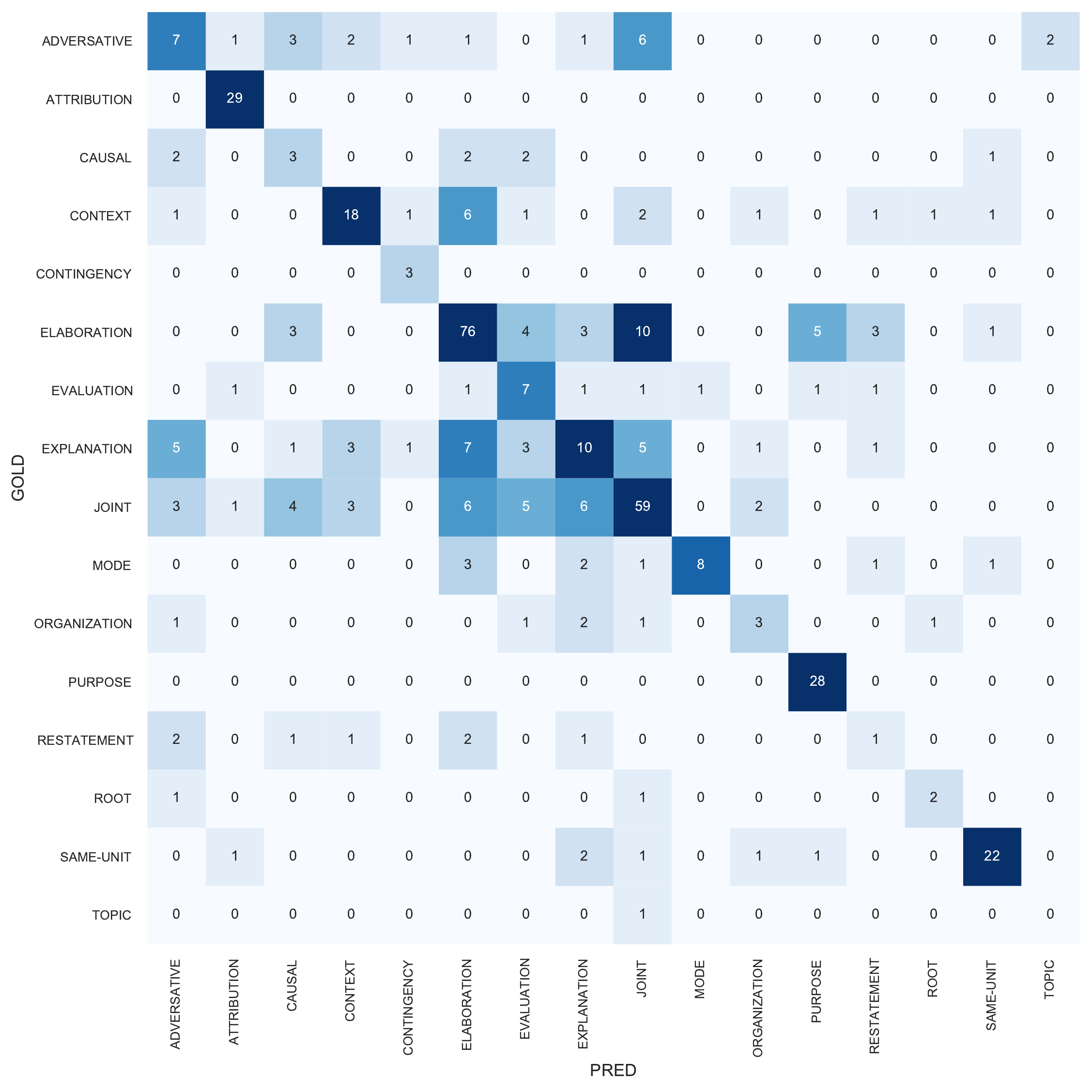}\label{fig:ova-speech}}
  \hfill 
  \subfloat[\textit{textbook}]{\includegraphics[width=70mm,scale=1]{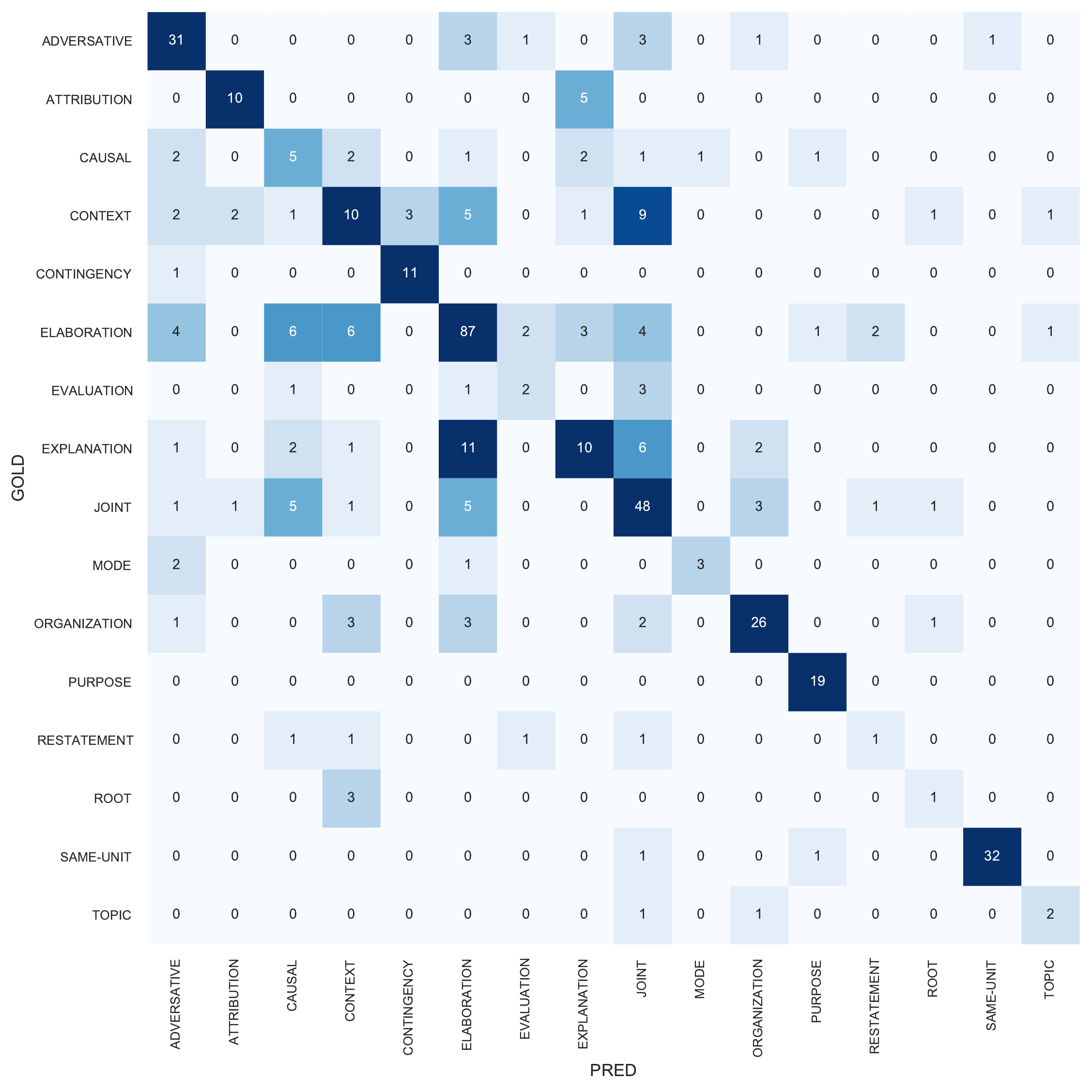}\label{fig:ova-textbook}}
  \hfill 
  \subfloat[\textit{vlog}]{\includegraphics[width=70mm,scale=1]{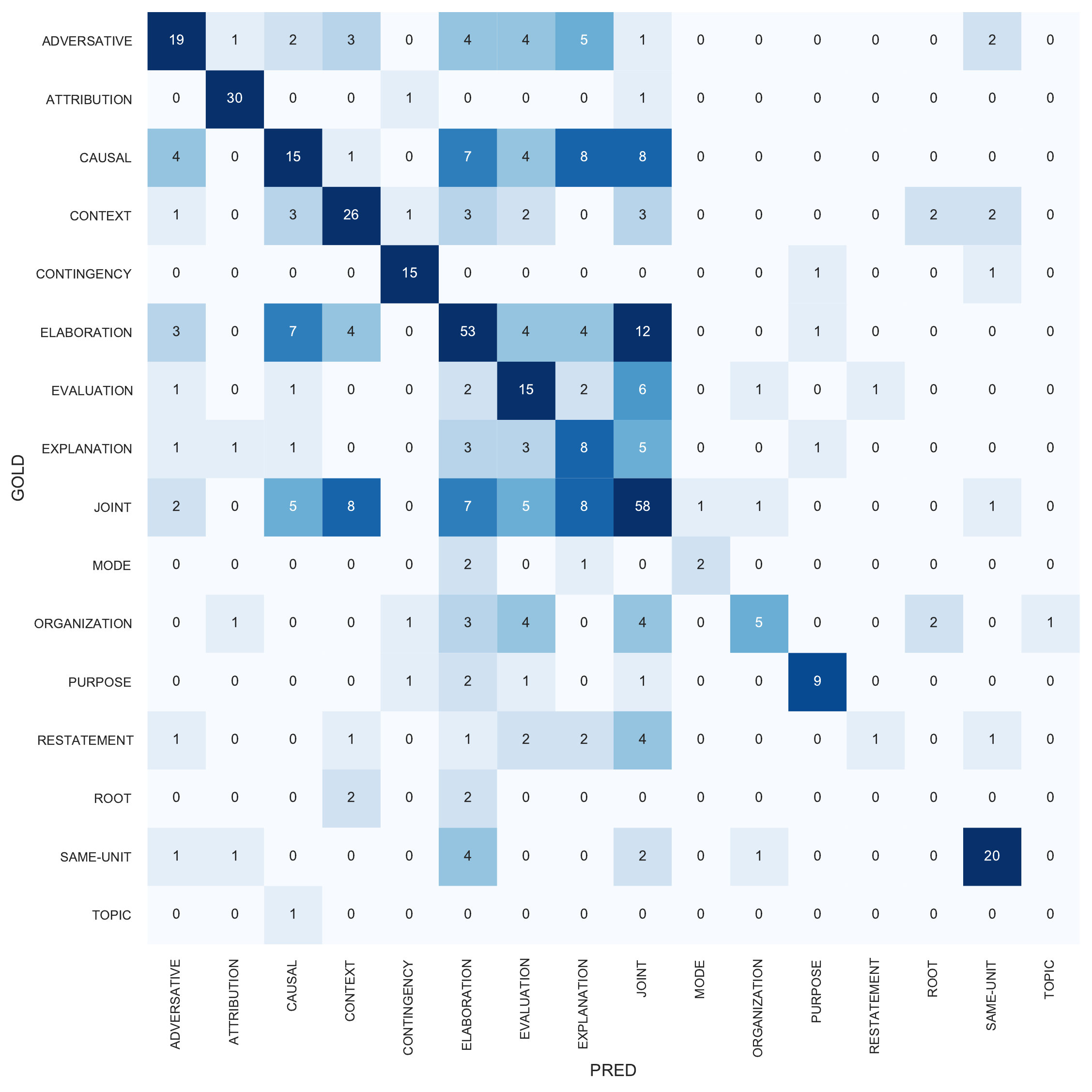}\label{fig:ova-vlog}} 
  
  \caption{Confusion Matrices for All Genres from their \textsc{ova} Models or the \textsc{all-large} Model from \S\ref{subsec:GUM8-cross-genre-ova}.}
\end{figure*}

\end{document}